\documentclass[11pt,a4paper]{article}
\usepackage[hyperref]{acl2021}

\usepackage{times}
\usepackage{todonotes}
\usepackage{latexsym}
\usepackage[T1]{fontenc}

\usepackage[utf8]{inputenc}

\usepackage{microtype}

\usepackage{amsmath,amsfonts,bm}

\def\eqref#1{equation~\ref{#1}}

\def\1{\bm{1}}

\DeclareMathAlphabet{\mathsfit}{\encodingdefault}{\sfdefault}{m}{sl}
\SetMathAlphabet{\mathsfit}{bold}{\encodingdefault}{\sfdefault}{bx}{n}

\usepackage{subcaption}
\usepackage{graphicx}
\usepackage{longtable}
\usepackage{multicol}
\usepackage{multirow}
\usepackage{algorithm}
\usepackage{supertabular}
\usepackage{microtype}
\usepackage{natbib}

\usepackage{color}

\usepackage{url}
\newtheorem{theorem}{Theorem}

\title{A Novel Estimator of Mutual Information for Learning to Disentangle Textual Representations}

\usepackage{algorithmicx}
\usepackage{algpseudocode}

\algnewcommand\algorithmicinput{\textbf{INPUT:}}
\algnewcommand\INPUT{\item[\algorithmicinput]}

\algnewcommand\algorithmicoutput{\textbf{OUTPUT:}}
\algnewcommand\OUTPUT{\item[\algorithmicoutput]}
\usepackage{changes}

\algnewcommand\algorithminit{\textbf{Initialization:}}
\algnewcommand\Initialize{\item[\algorithminit]}

\algnewcommand\algorithStandardization{\textbf{Threshold:}}
\algnewcommand\Standardization{\item[\algorithStandardization]}
\aclfinalcopy
\algnewcommand\algorithoptim{\textbf{Optimization:}}
\algnewcommand\Optimization{\item[\algorithoptim]}

\author{Pierre Colombo$^{\dag\star}$, Chlo\'e Clavel$^\star$, Pablo Piantanida$^*$ \\
  $^\star$T\'el\'ecom ParisTech, Universit\'e Paris Saclay \\
  $\dag$ IBM GBS France\\
  $^*$Laboratoire des Signaux et Systèmes (L2S), CentraleSupelec CNRS Universite Paris-Saclay \\
  \texttt{pierre.colombo@ibm.com} \\
  \texttt{chloe.clavel@telecom-paris.fr} \\
  \texttt{pablo.piantanida@centralesupelec.fr}
  }

\begin{document}

\maketitle
\begin{abstract}
Learning disentangled representations of textual data is essential for many natural language tasks such as fair classification, style transfer and sentence generation, among others. The existent dominant approaches in the context of text data {either rely} on training an adversary (discriminator) that aims at making  attribute values difficult to be inferred from the latent code {or rely on minimising variational bounds of the mutual information between latent code and the value attribute}. {However, the available methods suffer of the impossibility to provide 
a fine-grained control of the degree (or force) of disentanglement.} {In contrast to} {adversarial methods}, which are remarkably simple, although  the adversary seems to be performing perfectly well during the training phase, after it is completed a fair amount of information about the undesired  attribute still remains.  This paper introduces a novel variational upper bound to the mutual information between an attribute and the latent code of an encoder.  Our bound aims at controlling the approximation error via the Renyi's divergence, leading to both better disentangled representations and in particular, a precise control of the desirable degree of disentanglement {than state-of-the-art methods proposed for textual data}. Furthermore, it does not suffer from the degeneracy of other losses in multi-class scenarios. We show  the superiority of this method on fair classification and on textual style transfer tasks. Additionally, we provide new insights illustrating various trade-offs in style transfer when attempting to learn disentangled representations and quality of the generated sentence.
\end{abstract}

\section{Introduction}

Learning disentangled representations hold a central place to build rich embeddings of high-dimensional data. For a representation to be disentangled implies that it factorizes some latent cause or causes of variation as formulated by \cite{6472238}. For example, if there are two causes for the transformations in the data that do not generally happen together and are statistically distinguishable (e.g., factors occur independently), a maximally disentangled representation is expected to present a sparse structure that separates those causes. Disentangled representations have been shown to be useful for a large variety of data, such as video \cite{video}, image \cite{iclr_irrelevant}, text \cite{text}, audio \cite{audio}, among others, and applied to many different tasks, \emph{e.g.}, robust and fair classification \cite{adversarial_removal}, visual reasoning \cite{visual}, style transfer \cite{style_transfert_1}, conditional generation \cite{conditional_generation_1,conditional_generation}, few shot learning \cite{few_shot}, among others. 

In this work, we focus our attention on learning disentangled representations for text, as it remains overlooked by \cite{text}. Perhaps, one of the most popular applications of disentanglement in textual data is fair classification \cite{adversarial_removal,adversarial_removal_2} and sentence generation tasks such as style transfer \cite{text} or conditional sentence generation \cite{dt_info}. For fair classification, perfectly disentangled latent representations can be used to ensure fairness as the decisions are taken based on representations which are statistically independent from--or at least carrying limited information about--the protected attributes. However,  there exists a trade-offs between full disentangled representations and performances on the target task, as shown by~\cite{feutry2018learning}, among others. For sequence generation and in particular, for style transfer,  learning disentangled representations aim at allowing an easier transfer of the desired style. To the best of our knowledge, a depth study of the relationship between disentangled representations based {either} on adversarial losses solely {or} on $vCLUB-S$ and quality of the generated sentences remains overlooked. Most of the previous studies have been focusing on either trade-offs between metrics computed on the generated sentences \cite{iclr_emnlp} or performance evaluation of the disentanglement as part of (or convoluted with) more complex modules. This enhances the need to provide a fair evaluation of disentanglement methods  by isolating their individual contributions  \cite{iclr_workshop,dt_info}.\\
Methods to enforce disentangled representations {can be grouped into two different categories. The first category} relies on an adversarial term in the training objective that aims at ensuring that sensitive attribute values (\textit{e.g.} race, sex, style) as statistically independent as possible from the encoded latent representation. Interestingly enough, several works \cite{text,adversarial_removal,loss_1,loss_2,loss_3,loss_4,loss_5},  \citet{adversarial_removal} have recently shown that even though the adversary teacher seems to be performing remarkably  well during training, after the training phase, a fair amount of information about the sensitive attributes still remains, and can be extracted from the encoded representation. {The second category aim at minimising Mutual Information (MI) between encoded latent representation and the sensitive attribute values, \emph{i.e.},  without resorting to an adversarial discriminator}. MI acts as an universal measure of dependence since it captures non-linear and statistical dependencies of high orders between the involved quantities \cite{equitable}. However, estimating MI has been a long-standing challenge, in particular when dealing with high-dimensional data \cite{estimation, pichler2020estimation}. Recent methods rely on variational upper bounds. For instance, \cite{dt_info} study \texttt{vCLUB-S} \cite{cheng2020club} for sentence generation tasks. Although this approach improves on previous state-of-the-art methods, it does not allow to fine-tuning of the desired degree of disentanglement, i.e., it enforces light or strong levels of  disentanglement where only few features relevant to the input sentence remain (see \citet{feutry2018learning} for further discussion). 

\subsection{Our Contributions} 

We {develop} new tools to build disentangled textual representations and evaluate them on fair classification and two sentence generation tasks, namely, style transfer and conditional sentence generation. Our main contributions are summarized below: 
\begin{itemize}
    \item \textit{A novel objective to train disentangled representations from attributes.} To overcome some of the limitations of {both} adversarial losses {and $\texttt{vCLUB-S}$} we derive a novel upper bound to the MI which aims at correcting the approximation error via either the Kullback-Leibler \cite{kl} or Renyi \cite{renyi1961measures} divergences. This correction terms appears to be a key feature to fine-tuning the degree of disentanglement compared  to $\texttt{vCLUB-S}$.
    \item \textit{Applications and numerical results.} First, we demonstrate that the aforementioned surrogate is better suited than the widely used adversarial losses {as well as \texttt{vCLUB-S}} as it can provide better disentangled textual representations while allowing \emph{fine-tuning of the desired degree of disentanglement}. In particular, we show that our method offers a better accuracy versus disentanglement trade-offs for fair classification tasks.  We additionally demonstrate that our surrogate outperforms {both methods} when learning disentangled representations for style transfer and conditional sentence generation while not suffering (or degenerating) when the number of classes is greater than two, which is an apparent limitation of adversarial training. By isolating the disentanglement module, we identify and report existing trade-offs between different degree of disentanglement and quality of generated sentences. The later includes content preservation between input and generated sentences and accuracy on the generated style.
\end{itemize}

\section{Main Definitions and Related Works}
We introduce notations, tasks, and closely related work. Consider a training set $\mathcal{D} = \{(x_i,y_i)\}_{i=1}^n$ of $n$ sentences $x_i \in \mathcal{X}$ paired with  attribute values $y_i \in \mathcal{Y}\equiv \{1,\dots, |\mathcal{Y}|\}$ which indicates a discrete attribute to be disentangled from the resulting representations. We study the following scenarios: 

\textbf{Disentangled representations.}  Learning disentangled representations consists in learning a model $\mathcal{M} : \mathcal{X} \rightarrow \mathcal{R}^d$ that maps feature inputs $X$ to a vector of dimension $d$ that retains as much as possible information of the original content from the input sentence but as little as possible about the undesired attribute $Y$. In this framework, content is defined as any relevant information present in $X$ that does not depend on $Y$.



\textbf{Applications to binary fair classification.} The task of fair classification through disentangled representations aims at building representations that are independent of selective discrete (sensitive) attributes (\textit{e.g.}, gender or race). This task consists in learning a model $\mathcal{M} : \mathcal{X} \rightarrow \{0,1\}$ that maps any input $x$ to a label $l\in \{0,1\}$. The goal of the learner is to build a predictor that assigns each $x$ to either $0$ or $1$ ``oblivious'' of the protected attribute $y$.  Recently, much progress has been made on devising appropriate means of fairness, \emph{e.g.}, \cite{fair_1,fair_2,fair_3}.
In particular,  \cite{adv_classif_fair_1,adversarial_removal_2,adversarial_removal} approach the problem based on adversarial losses. More precisely, these approaches consist in learning an encoder that maps $x$ into a representation vector $h_{x}$, a critic $C_{\theta_c}$ which attempts to predict $y$, and an output classifier $f_{\theta_d}$ used to predict $l$ based on the observed $h_{x}$. The classifier is said to be fair if there is no statistical information about $y$ that is present in $h_{x}$ \cite{adv_classif_fair_1,adversarial_removal}.

\textbf{Applications to conditional sentence generation.} The task of conditional sentence generation consists in taking an input text containing specific stylistic properties to then generate a realistic (synthetic) text containing potentially different stylistic properties. It requests to learn a model $\mathcal{M} : \mathcal{X} \times  \mathcal{Y}  \rightarrow \mathcal{X}$ that maps a pair of inputs $(x,y^t)$ to a sentence $x^g$, where the outcome sentence should retain as much as possible of the original content from the input sentence while having (potentially a new) attribute $y^g$. Proposed approaches to tackle textual style transfer \cite{formality_1,formality} can be divided into two main categories. The first category \cite{back,multiple} uses cycle losses based on back translation \cite{back_translation_paper} to ensure that the content is preserved during the transformation. Whereas, the second category {look to explicitly separate attributes from the content. This constraint is enforced using either} adversarial training \cite{style_transfert_1,loss_5,loss_4,iclr_workshop} {or MI minimisation using \texttt{vCLUB-S} \cite{dt_info}}. Traditional {adversarial} training is based on an encoder that aims to fool the adversary discriminator by removing attribute information from the content embedding \cite{adversarial_removal}. As we will observe, the more the representations are  disentangled the easier is to transfer the style but at the same time the less the content is preserved. In order to approach the sequence generation tasks, we build on the Style-embedding Model by  \cite{text} (StyleEmb) which uses adversarial losses introduced in prior work for these dedicated tasks. During the training phase, the input sentence is fed to a sentence encoder, namely $f_{\theta_e}$, while the input style is fed to a separated style encoder, namely $f_{\theta_e}^s$. During the inference phase, the desired style--potentially different from the input style--is provided as input along with the input sentence. 
\section{Model and Training Objective}
This section describes the proposed approach to learn disentangled representations. We first review MI along with the model overview  and then, we derive the variational bound we will use, and discuss connections with adversarial losses.

\subsection{Model Overview}\label{ssec:model_overview}
The MI is a key concept in information theory for measuring high-order statistical  dependencies between random quantities. Given two random variables $Z$ and $Y$, the MI is defined by
\begin{equation}\label{eq:mi_def}
    I(Z;Y) = \mathop{\mathbb{E}_{ZY}} \left[ \log \frac{p_{ZY}(Z,Y)}{p_Z(Z)p_Y(Y)} \right], 
\end{equation}
where $p_{ZY}$ is the joint probability density function (pdf) of the random variables $(Z,Y)$, with $p_Z$ and $p_Y$ representing the respective marginal pdfs.
MI is related to entropy $h(Y)$ and conditional   entropy  $h(Y|Z)$ as  follows:
\begin{equation}
    I(Z;Y) = h(Y) - h(Y|Z). 
\end{equation}
Our models for fair classification and sequence generation share a similar structure. These rely on an encoder that takes as input a random sentence $X$ and maps it to a random representation $Z$ using a deep encoder denoted by $f_{\theta_e}$. Then, classification and sentence generation are performed using either a classifier or an auto-regressive decoder denoted by $f_{\theta_d}$. We aim at minimizing MI between the latent code represented by the Random Variable (RV) $Z=f_{\theta_e}(X)$ and the desired attribute represented by the RV $Y$. The objective of interest $\mathcal{L}(f_{\theta_e})$ is defined as:
\begin{equation}\label{eq:all_loss}
    \mathcal{L}(f_{\theta_e}) \equiv  \underbrace{\mathcal{L}_{down.}(f_{\theta_e})}_{\textrm{downstream task}} + \lambda \cdot \underbrace{I(f_{\theta_e}(X);Y)}_{\textrm{disentangled}},
\end{equation}
where $\mathcal{L}_{down.}$ represents a downstream  specific (target task) loss and $\lambda$ is a meta-parameter that controls the sensitive trade-off between disentanglement (\emph{i.e.}, minimizing MI) and success in the downstream task (\emph{i.e.}, minimizing the target loss). In \autoref{sec:numerical_result}, we illustrate theses different trade-offs.

\textbf{Applications to fair classification and sentence generation.} For fair classification, we follow standard practices and optimize the cross-entropy between prediction and ground-truth labels. 
In the sentence generation task $\mathcal{L}_{down.}$ represents the  negative log-likelihood between individual tokens. 

\subsection{A Novel Upper Bound on MI}\label{ssec:bounds}
Estimating the MI is a long-standing challenge as the exact computation \cite{estimation} is only tractable for discrete variables, or for a limited family of problems where the underlying data-distribution satisfies smoothing properties, see recent work by \cite{pichler2020estimation}. Different from previous approaches leading to variational lower bounds \cite{mine,evo_mine,nce}, in this paper we derive an estimator  based on a  variational upper bound to the MI which control the approximation error  based on the Kullback-Leibler and the Renyi divergences \cite{daudel:hal-02614605}.  

\begin{theorem}\label{th:one}(Variational upper bound on MI) 
Let $(Z, Y)$ be an arbitrary pair of RVs with $(Z, Y) \sim p_{ZY}$ according to some underlying  pdf, and let $q_{\widehat{Y}|Z}$ be a conditional variational distribution on the attributes  satisfying $p_{ZY} \ll p_Z\cdot  q_{\widehat{Y}|Z}$, i.e., absolutely continuous. Then, we have that
\begin{align}\label{eq:kl_bound1}
    \begin{split}
        &I(Z;Y) \leq \mathbb{E}_{Y} \left[-\log\int q_{\widehat{Y}|Z}(Y|z) p_Z(z)  dz \right] + \\ & \mathbb{E}_{YZ } \left[\log q_{\widehat{Y}|Z}(Y|Z)\right] + \text{KL}\big(p_{ZY}\| p_Z \cdot q_{\widehat{Y}|Z}\big),  
        \end{split}
\end{align}
where $\textrm{KL}\big(p_{ZY}\| p_Z \cdot q_{\widehat{Y}|Z}\big)$ denotes the KL divergence. Similarly, we have that for any $\alpha > 1$,
\begin{align}\label{eq:renyi_bound2}
    \begin{split}
&I(Z;Y) \leq  \mathbb{E}_{Y} \left[-\log\int  q_{\widehat{Y}|Z}(Y|z) p_Z(z) dz \right] + \\& \mathbb{E}_{YZ } \left[\log q_{\widehat{Y}|Z}(Y|Z)\right] +  D_\alpha\big(p_{ZY}\| p_Z \cdot q_{\widehat{Y}|Z}\big), 
\end{split}
\end{align}
where $(\alpha - 1)D_\alpha\big(p_{ZY}\| p_Z \cdot q_{\widehat{Y}|Z}\big) =  \log \mathbb{E}_{ZY}[ $ $R^{\alpha-1}(Z,Y)]$ 
denotes the Renyi divergence and $R(z, y) = \frac{p_{Y|Z}(y|z)}{q_{\widehat{Y}|Z}(y|z)}$, for  $(z,y)\in\textrm{Supp}(p_{ZY})$.
\end{theorem}

\textit{Proof:} The upper bound on $H(Y)$ is a direct application of the the \cite{donsker1985large} representation of KL divergence while the lower bound on $H(Y|Z)$ follows from the monotonicity property of the function: $\alpha \mapsto D_\alpha\big(p_{ZY}\| p_Z \cdot q_{\widehat{Y}|Z}\big)$. Further details are relegated to \autoref{sec:proofs}.

\textbf{Remark:} It is worth to emphasise that the KL divergence in \eqref{eq:kl_bound1} and Renyi divergence in \eqref{eq:renyi_bound2} control the approximation error between the exact entropy and its corresponding bound.


\textbf{From theoretical bounds to trainable surrogates to minimize MI:} It is easy to check that the inequalities in (\autoref{eq:kl_bound1}) and (\autoref{eq:renyi_bound2}) are tight provided that $p_{ZY}\equiv  p_Z\cdot  q_{\widehat{Y}|Z}$ \emph{almost surely} for some adequate choice of the variational distribution. However, the evaluation of these bounds requires to obtain an estimate of the density-ratio $R(z, y)$. Density-ratio estimation has been widely studied in the literature (see \cite{10.5555/2181148} and references therein) and confidence bounds has been reported by \cite{pmlr-v54-kpotufe17a} under some smoothing assumption on underlying data-distribution $p_{ZY}$. In this work, we will estimate this ratio by using a critic $C_{\theta_R}$ which is trained to differentiate between a balanced dataset of positive i.i.d samples coming from $p_{ZY}$ and negative i.i.d samples coming from $q_{\widehat{Y}|Z}\cdot p_{Z}$. Then, for any pair $(z,y)$, the density-ratio can be estimated by 
$R(z, y) \approx \frac{ \sigma(C_{\theta_R}(z,y))}{1 - \sigma(C_{\theta_R}(z,y))}$, where $\sigma(\cdot)$ indicates the sigmoid function and $C_{\theta_R}(z,y)$ is the unnormalized output of the critic.  It is worth to mention that after estimating this ratio, the previous upper bounds may not be strict bounds so we will refer them as surrogates. 

\subsection{Comparison to {existing methods}}\label{ssec:adv_loss}
{\textbf{Adversarial approaches}:} In order to enhance our understanding of why the proposed approach based on the minimization of the MI using our variational upper bound in \autoref{th:one} may lead to a better training objective than previous  adversarial losses, we discuss below the explicit relationship between MI and cross-entropy loss. Let $Y\in\mathcal{Y}$ denote a random attribute and let $Z$ be a possibly high-dimensional representation  that needs to be disentangled from $Y$.  Then,  
\begin{align}\label{eq:adv_losses_mi}
    \begin{split}
            I(Z;Y) \,\geq \, & H(Y) -  \mathbb{E}_{YZ } \left[\log q_{\widehat{Y}|Z}(Y|Z)\right] \\ \, = \, & \textrm{Const} - \textrm{CE}(\widehat{Y}|Z),
    \end{split}
\end{align}
where $\textrm{CE}(\widehat{Y}|Z)$ denotes the cross-entropy corresponding to the  adversarial discriminator $q_{\widehat{Y}|Z}$, noting that $Y$ comes from an unknown distribution on which we have no influence $H(Y)$ is an unknown constant, and using that the approximation error:  $\textrm{KL}\big(q_{ZY}\|q_{\widehat{Y}|Z} \cdot p_Z\big)  = \textrm{CE}(\widehat{Y}|Z) - H(Y|Z)$. \autoref{eq:adv_losses_mi} shows that the cross-entropy loss leads to a lower bound  (up to a constant) on the MI. Although  the cross-entropy can lead to good estimates of the conditional entropy, the adversarial approaches for classification and sequence generation by \cite{adversarial_removal_2,text} which consists in maximizing the cross-entropy, induces a degeneracy (unbounded loss)  as $\lambda$ increases in the underlying optimization problem. As we will observe in next section, our variational upper bound in \autoref{th:one} can overcome this issue, in particular for $|\mathcal{Y}|>2$. 

{\textbf{\texttt{vCLUB-S}}: Different from our method, \citet{cheng2020club} introduce $I_{\text{vCLUB}}$ which is an upper bound on MI defined by 
\begin{align}\label{eq:club}
\begin{split}
I_{\text{vCLUB}}(Y;Z) = & \mathbb{E}_{YZ}[\log p_{Y|Z}(Y|Z)] \\ & - \mathbb{E}_{Y}\mathbb{E}_{Z}[\log p_{Y|Z}(Y|Z)].
\end{split}
\end{align}
It would be worth to mention that this bound follows a similar approach to the  previously introduced bound in \cite{feutry2018learning}.}
\section{Experimental Setting}

\subsection{Datasets}\label{ssec:datasets}
\textbf{Fair classification task.} We follow the experimental protocol of \cite{adversarial_removal}. The main task consists in predicting a binary label representing either the sentiment (positive/negative) or the mention. The mention task aims at predicting if a tweet is conversational. Here the considered protected attribute is the race. The dataset has been automatically constructed from DIAL corpus \cite{dial} which contained race annotations over $50$ Million of tweets. Sentiment tweets are extracted using a list of predefined emojis and mentions are identified using @mentions tokens. The final dataset contains 160k tweets for the training and two splits of 10K tweets for validation and testing. Splits are balanced such that the random estimator is likely to achieve $50\%$ accuracy.
\textbf{Style Transfer} For our sentence generation task, we conduct experiments on three different datasets extracted from restaurant reviews in Yelp. The first dataset, referred to as SYelp, contains 444101, 63483, and 126670 labelled short reviews (at most 20 words) for train, validation, and test, respectively. 
For each review a binary label is assigned depending on its polarity. 
Following \cite{multiple}, we use a second version of Yelp, referred to as FYelp, with longer reviews (at most $70$ words). It contains 
five coarse-grained restaurant category labels (\textit{e.g.}, Asian, American, Mexican, Bars and Dessert). The multi-category FYelp is used to access the generalization capabilities of our methods to a multi-class scenario. 

\subsection{Metrics for Performance Evaluation}\label{sec:metric} 
\textbf{Efficiency measure of the disentanglement methods.} \cite{adversarial_removal_2} report that offline classifiers (post training) outperform clearly adversarial discriminators.  We will re-training a classifier on the latent representation learnt by the model and we will report its accuracy. 

\textbf{Measure of performance within the fair classification task.} In the fair classification task we aim at maximizing accuracy on the target task and so we will report the corresponding accuracy. 

\textbf{Measure of performance within sentence generation tasks.} Sentences generated by the model are expected to be fluent, to preserve the input content and to contain the desired style. For style transfer, the desired style is different from the input style while for conditional sentence generation, both input and output styles should be similar. Nevertheless, automatic evaluation of generative models for text is still an open problem. We measure the style of the output sentence by using a fastText classifier \cite{fastext_1}. For content preservation, we follow \cite{text} and compute both: (i) the cosine measure between source and generated sentence embeddings, which are the concatenation of min, max, and mean of word embedding (sentiment words removed), and (ii) the BLEU score between generated text and the input using SACREBLEU from \cite{sacrebleu}. Motivated by previous work, we evaluate the fluency of the language with the perplexity given by a GPT-2 \cite{gpt} pretrained model performing fine-tuning on the training corpus. We choose to report the log-perplexity since we believe it can better reflects the uncertainty of the language model (a small variation in the model loss would induce a large change in the perplexity due to the exponential term). Besides the automatic evaluation, we further test our disentangled representation effectiveness by human evaluation results are presented in \autoref{tab:human_annotation}.
\\\textbf{Conventions and abbreviations.} $Adv$ refers to a model trained using the adversarial loss; {\texttt{vCLUB-S}, $\textrm{KL}$ refers to a model trained using the \texttt{vCLUB-S} and KL surrogate (see \autoref{eq:kl_bound}) respectively}; and $D_\alpha$ refers to a model trained based on the $\alpha$-Renyi surrogate (\autoref{eq:renyi_bound}), for $\alpha\in\{1.3,1.5,1.8\}$.
\section{Numerical Results}\label{sec:numerical_result}
In this section, we present our results  on the fair classification  and binary sequence generation tasks, see \autoref{ssec:fairness} and  \autoref{ssec:bsg}, respectively. We additionally show that our variational surrogates to the MI--contrarily to adversarial losses--do not suffer in multi-class scenarios (see \autoref{ssec:msg}).

\subsection{Applications to Fairness}\label{ssec:fairness}
\begin{figure*} \centering
    \begin{subfigure}[t]{0.43\textwidth}
        \includegraphics[width=\textwidth]{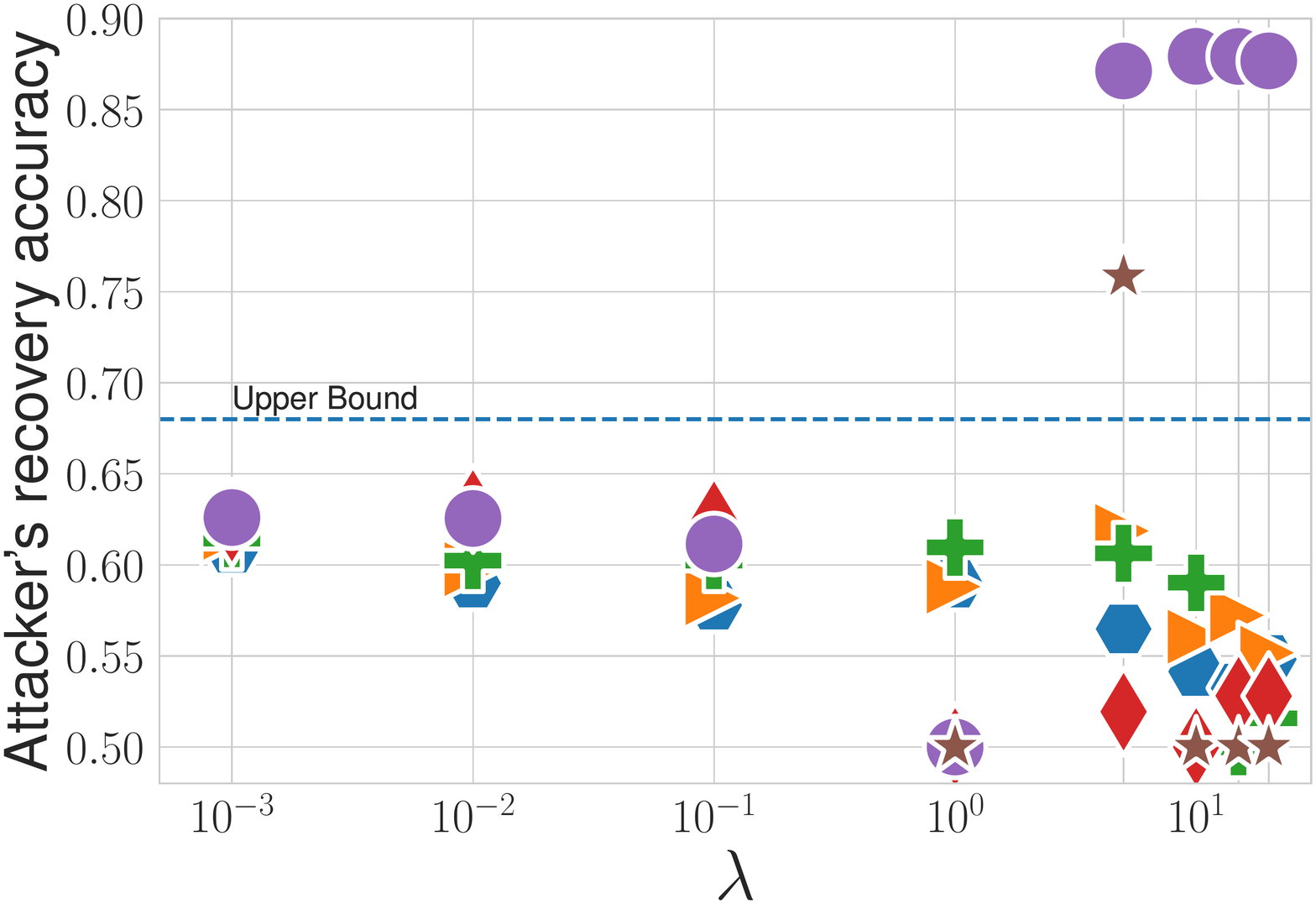}
         \vspace{-.6cm}
         \caption{}\label{fig:attacker_accuracy_mention}
    \end{subfigure}
    \begin{subfigure}[t]{0.43\textwidth} 
        \includegraphics[width=\textwidth]{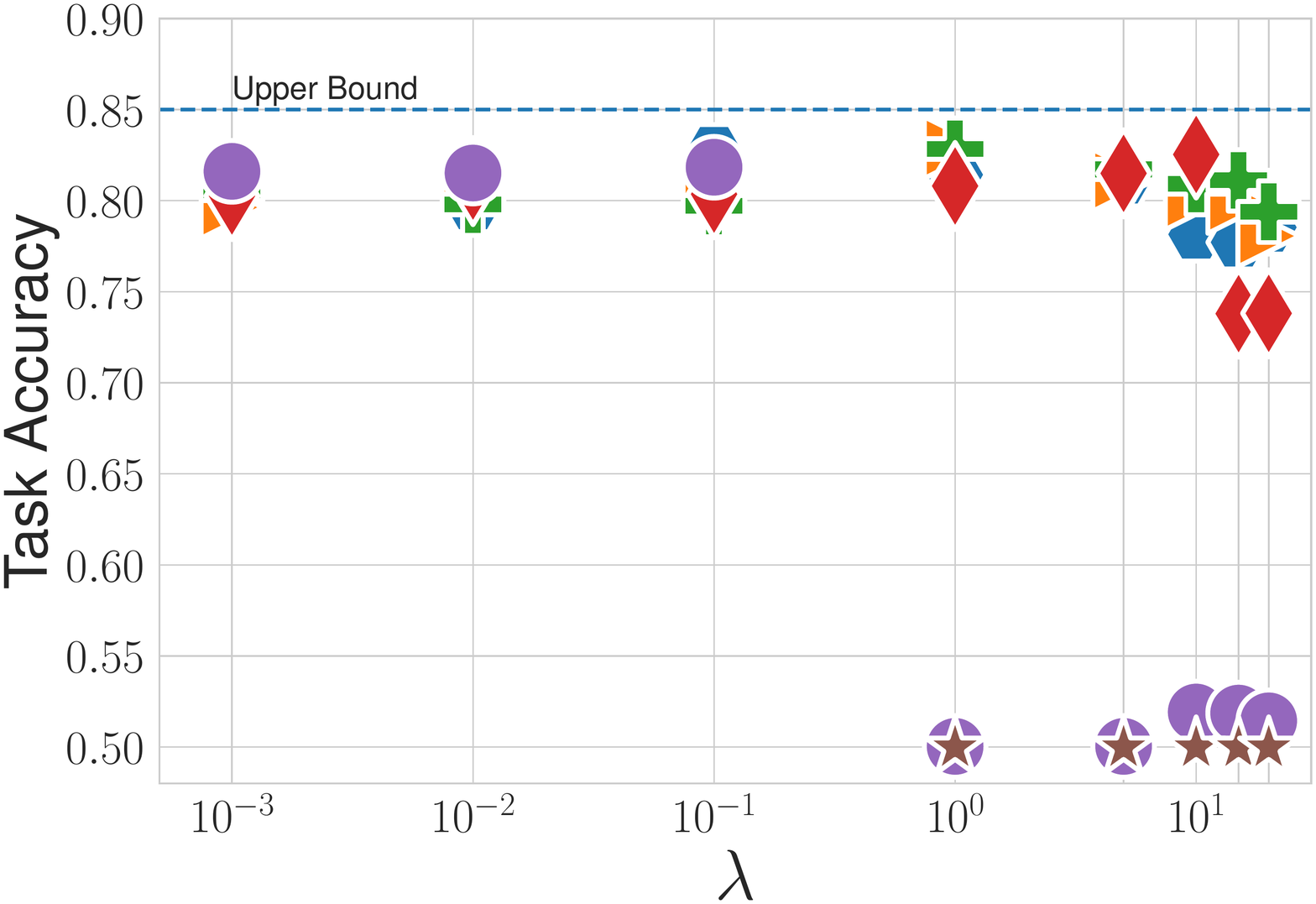}
         \vspace{-.6cm}
         \caption{}\label{fig:task_accuracy_mention}
    \end{subfigure} 
    \centering
        \begin{subfigure}[t]{0.43\textwidth}
        \includegraphics[width=\textwidth]{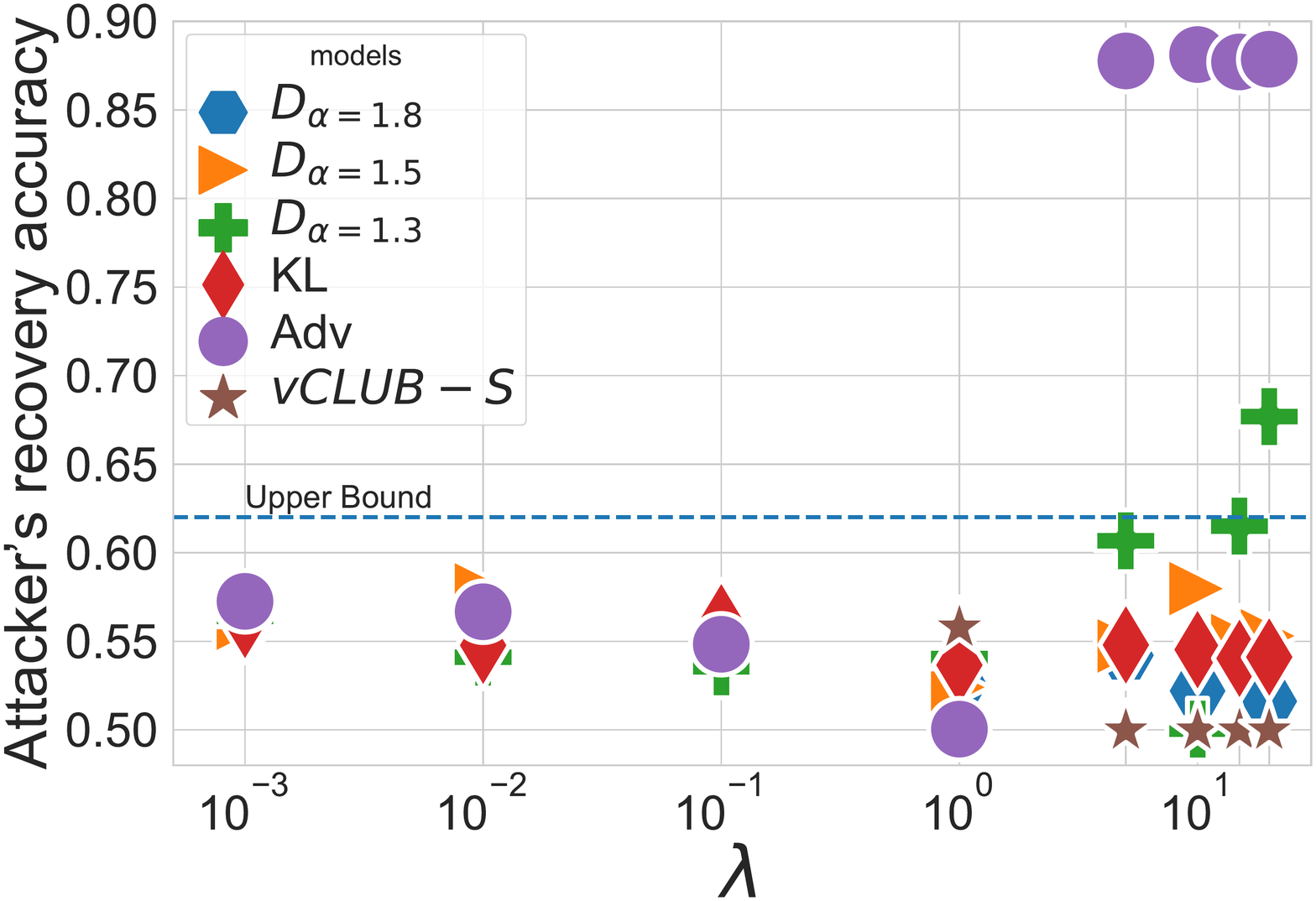}         \vspace{-.6cm}
         \caption{}\label{fig:attacker_accuracy_sentiment}         \vspace{-.6cm}
    \end{subfigure}
    \begin{subfigure}[t]{0.43\textwidth}
        \includegraphics[width=\textwidth]{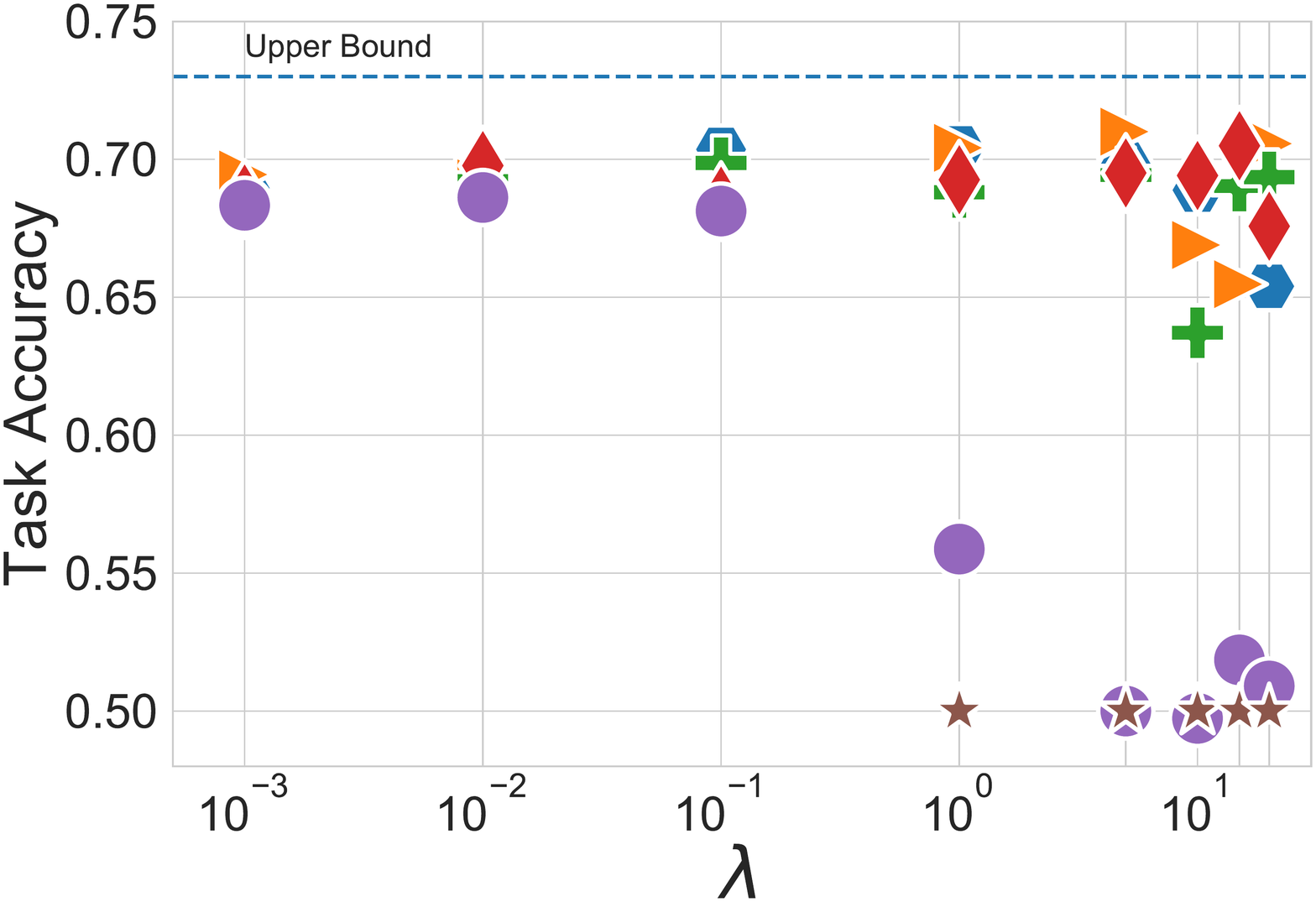}         \vspace{-.6cm}
         \caption{}\label{fig:task_accuracy_sentiment}          \vspace{-.5cm}
    \end{subfigure}
        \caption{Numerical results on fair classification. Trade-offs between target task and attacker accuracy are reported in \autoref{fig:attacker_accuracy_mention},  \autoref{fig:task_accuracy_mention} for mention task, and  \autoref{fig:attacker_accuracy_sentiment}, \autoref{fig:task_accuracy_sentiment} for sentiment task. For low values of $\lambda$ some points coincide. As $\lambda$ increases the level of disentanglement increases and the proposed methods using both KL ($KL$) and Reny divergences ($\mathcal{D}_\alpha$) clearly offer better control than existing methods. }\label{fig:all_fair_classification}
\end{figure*}

\textbf{Upper bound on performances.} We first examine how much of the protected attribute we can be recovered from an unfair classifier (\textit{i.e.}, trained without adversarial loss) and how well does such classifier perform. Results are reported in \autoref{fig:all_fair_classification}. We observe that we achieve similar scores than the ones reported in previous studies \cite{adversarial_removal_2,adversarial_removal}. This experiment shows that, when training to solve the main task, the classifier learns information about the protected attribute, \emph{i.e.}, the attacker's accuracy is better than random guessing. In the following, we compare the different proposed methods to disentangle representations and obtain a fairer classifier.

\textbf{Methods comparisons.} \autoref{fig:all_fair_classification} shows the results of the different models and illustrates the trade-offs between disentangled representations and the target task accuracy. Results are reported on the testset for both  sentiment and mention tasks when race is the protected. We observe that the classifier trained with an adversarial loss degenerates for $\lambda > 5$ since the adversarial term in \autoref{eq:all_loss} is influencing much the global gradient than the downstream term (\textit{i.e.}, cross-entropy loss between predicted and golden distribution). Remarkably, both models trained to minimize either the KL or the Renyi surrogate do not suffer much from the aforementioned multi-class problem. 
For both tasks, we observe that the KL and the Renyi surrogates can offer better disentangled representations than those induced by adversarial approaches. In this task, both the KL and Renyi achieve perfect  disentangled representations (\emph{i.e.}, random guessing accuracy on protected  attributes) with a $5\%$ drop in the accuracy of the target task,  when perfectly masking the protected attributes. {As a matter of fact, we observe that $\texttt{vCLUB-S}$ provides only two regimes: either a ``light'' protection (attacker accuracy around 60\%), with almost no loss in task accuracy ($\lambda < 1$), or a strong protection (attacker accuracy around 50\%), where a few features relevant to the target task remain.}\footnote{This phenomenon is also reported in \cite{feutry2018learning} on a picture anonymization task.} On the sentiment task, we can draw similar conclusions. However, the Renyi's surrogate achieves slightly better-disentangled representations. Overall, we can observe that our proposed surrogate enables good control of the degree of  disentangling. Additionally, we do not observe a degenerated behaviour--as it is the case with adversarial losses--when $\lambda$ increases. Furthermore, our surrogate allows simultaneously better disentangled representations while preserving the accuracy of the target task.  

\subsection{Applications to binary polarity transfer}\label{ssec:bsg}
In the previous section, we have shown that the proposed surrogates do not suffer from limitations of adversarial losses and allow to achieve better disentangled representations {than existing methods relying on \texttt{vCLUB-S}}. {Disentanglement modules} are a core block for a large number of both style transfer and conditional sentence generation algorithms \cite{iclr_emnlp,iclr_workshop,style_transfert_1} that place explicit constraints to force disentangled representations. First, we assess the disentanglement quality and the control over desired level of disentanglement while changing the downstream term, which for the sentence generation task is the cross-entropy loss on individual token. Then, we exhibit the existing trade-offs between quality of generated sentences, measured by the metric introduced in \autoref{sec:metric},  and the resulting degree of disentanglement.
The results are presented for SYelp
\subsubsection{Evaluating disentanglement} 
\begin{figure*}
\centering    \begin{subfigure}[t]{0.46\textwidth}
    \includegraphics[width=\textwidth]{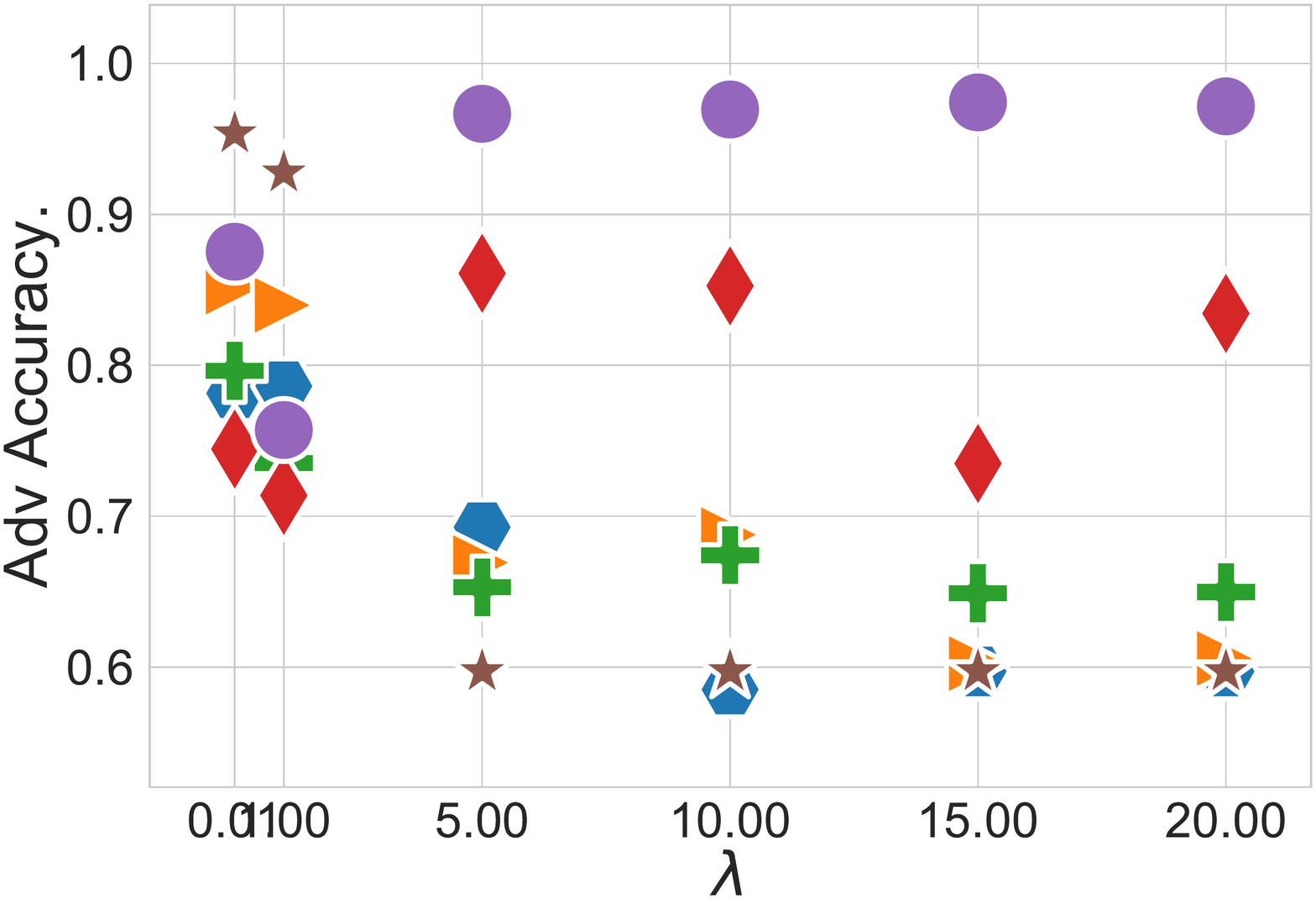}\vspace{-4mm}
      \caption{Binary Style  Transfert.}\label{fig:disent_sentiment_st}\vspace{-.5cm}
\end{subfigure}\begin{subfigure}[t]{0.46\textwidth}
    \centering
    \includegraphics[width=\textwidth]{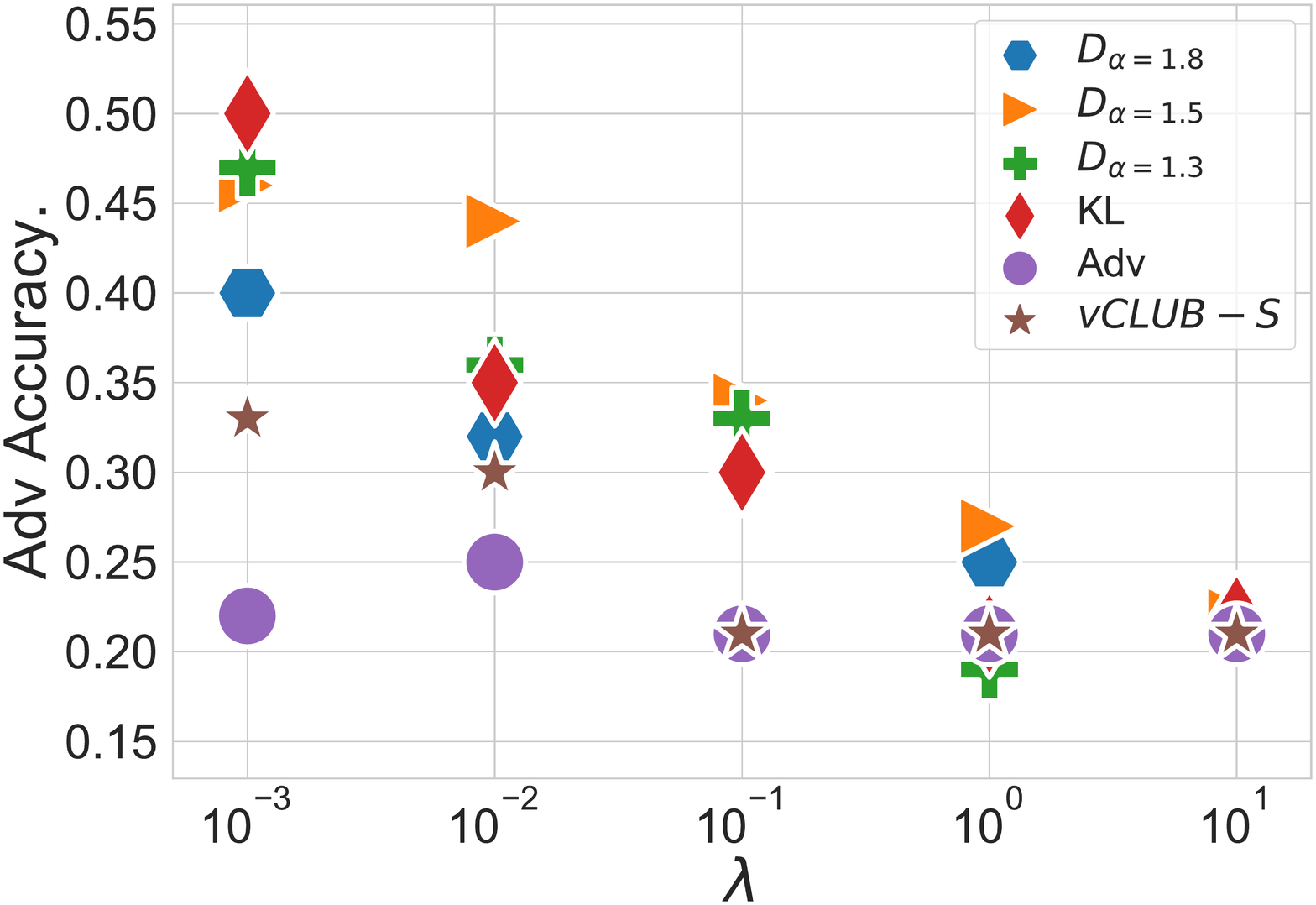}\vspace{-4mm}
    \caption{Multiclass Style Transfert}
\label{fig:disent_sentiment_mst}\vspace{-.3cm}
     \end{subfigure}      \caption{Disentanglement of representation learnt by $f_{\theta_e}$ in the binary (left) and multi-class (\textit{i.e.}, $|\mathcal{Y}|=5$) (right) sentence generation scenario. In the multi-class scenario the $Adv$ degenerates for $\lambda \geq0.01$ and offer no fined-grained control over the degree of disentanglement.}
\end{figure*}

\autoref{fig:disent_sentiment_st} shows the adversary accuracy of the different methods as a function of $\lambda$. Similarly to the fair classification task, a fair amount of information can be recovered from the embedding learnt with adversarial loss. In addition, we observe a clear degradation of its performance for values $\lambda > 1$. In this setting, the Renyi surrogates achieves consistently better results in terms of disentanglement than the one minimizing the KL surrogate. The curve for Renyi's surrogates shows that exploring different values of $\lambda$ allows good control of the disentanglement degree. Renyi surrogate generalizes well for sentence generation. {Similarly to the fairness task vCLUB-S only offers two regimes: "light" disentanglement with very little polarity transfer and "strong" disentanglement.} 

\begin{figure*} 
\centering    \hspace{-0.5cm}\begin{subfigure}[t]{0.34\textwidth}
        \includegraphics[width=\textwidth]{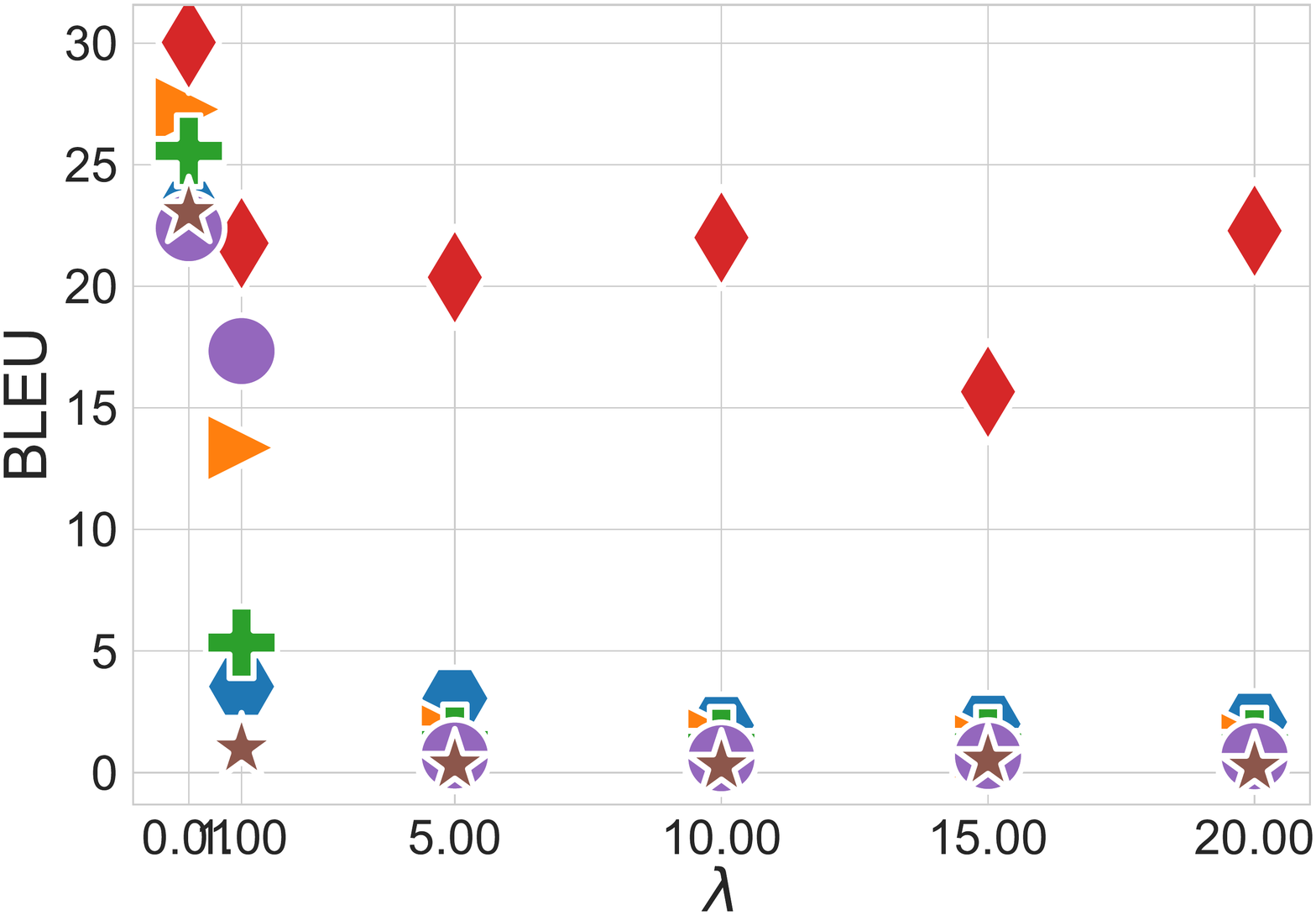}\vspace{-.3cm}
        \caption{}\label{fig:bleu_st}  
    \end{subfigure} 
\begin{subfigure}[t]{0.34\textwidth}
        \includegraphics[width=\textwidth]{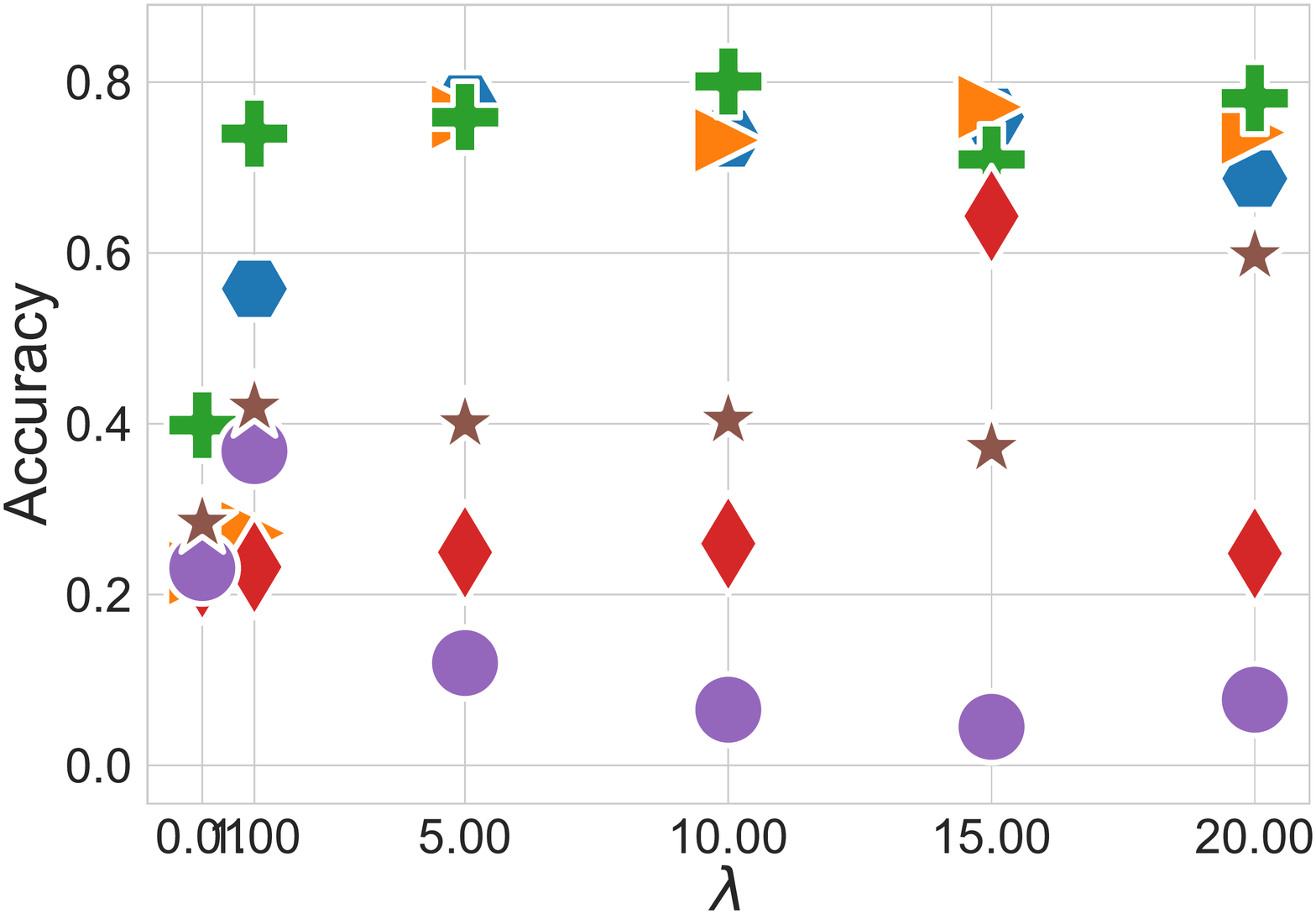}\vspace{-.3cm}
        \caption{}\label{fig:accuracy_sentiment_st} 
    \end{subfigure} \begin{subfigure}[t]{0.34\textwidth}
        \includegraphics[width=\textwidth]{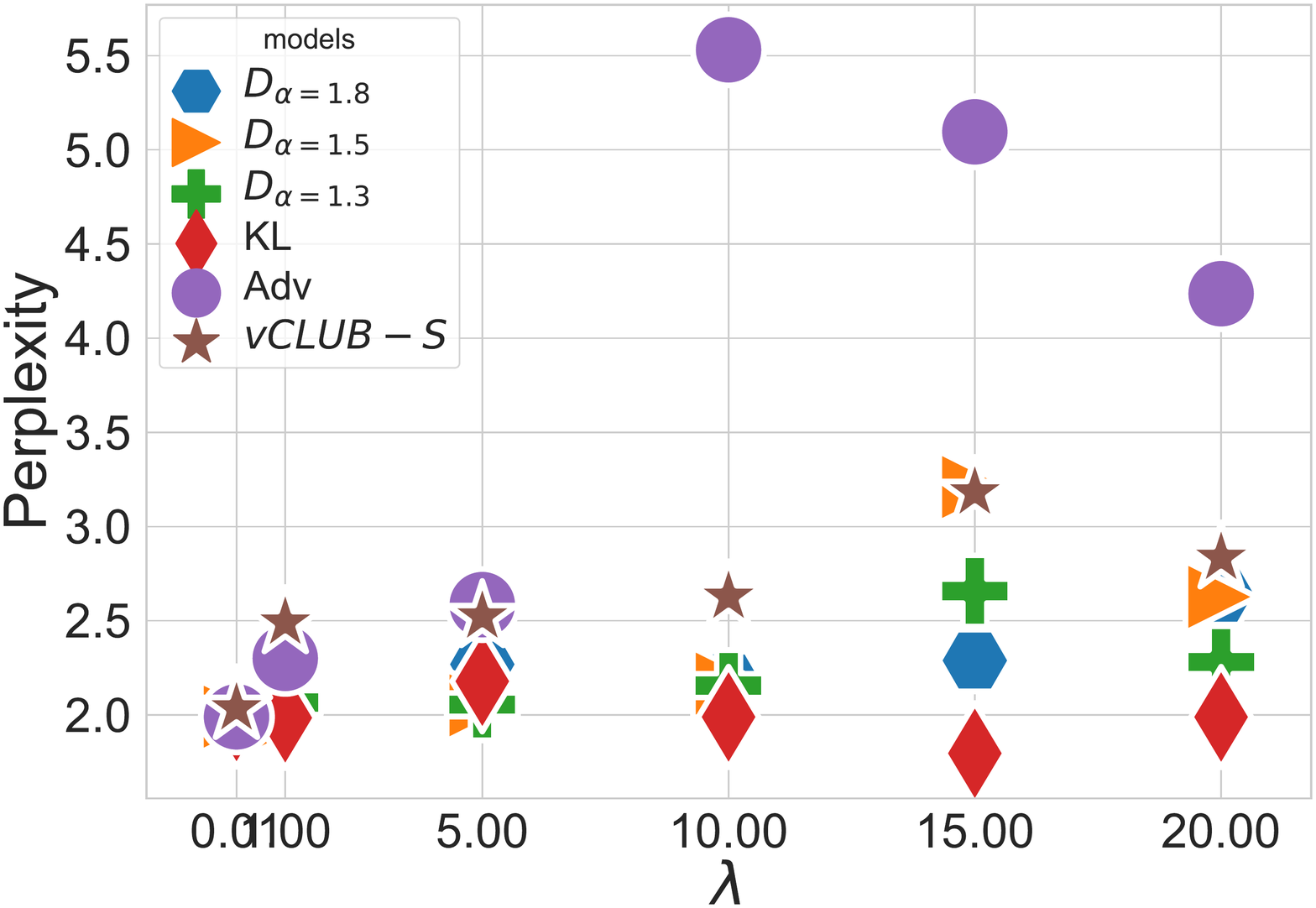}\vspace{-.3cm}
         \caption{}\label{fig:ppl_sentiment_st}
    \end{subfigure}\vspace{-.3cm}
        \caption{Numerical experiments on binary style transfer. Quality of generated sentences are evaluated using BLEU (\autoref{fig:bleu_st}); style transfer accuracy (\autoref{fig:bleu_st}); sentence fluency  (\autoref{fig:ppl_sentiment_st}). We report existing trade-offs between disentanglement and sentence generation quality. Human evaluation is reported in \autoref{tab:human_annotation}. }\label{fig:style_transfert_sentiment}\vspace{-.3cm}

\end{figure*}

\subsubsection{Disentanglement in Polarity Transfer}
The quality of generated sentences are evaluated using the fluency (see \autoref{fig:ppl_sentiment_st} 
), the content preservation (see \autoref{fig:bleu_st})
, additional results using a cosine similarity are given in \autoref{sec:addition_sentiment}, and polarity accuracy (see \autoref{fig:accuracy_sentiment_st} 
). For style transfer, and for all models, we observe trade-offs  between disentanglement and content preservation (measured by BLEU) and between fluency and disentanglement. Learning disentangled representations leads to poorer content preservation. As a matter of fact, similar conclusions can be drawn while measuring content with the cosine similarity (see \autoref{sec:addition_sentiment}). For polarity accuracy, in non-degenerated cases (see below), we observe that the model is able to better transfer the sentiment in presence of disentangled representations. \textit{Transferring style is easier with disentangled representations, however there is no free lunch here since disentangling also removes important information about the content}. {It is worth noting that even in the "strong" disentanglement regime \texttt{vCLUB-S} struggles to transfer the polarity (accuracy of 40\% for $\lambda \in \{1,2,10,15\}$) where other models reach 80\%.} 
It is worth noting that similar conclusions hold for two different sentence generation tasks: style transfer and conditional generation, which tends to validate the current line of work that formulates text generation as generic text-to-text \cite{t2t}. 
\\\textbf{Quality of generated sentences.} Examples of generated sentences are given in \autoref{tab:example_sentences_st} 
, providing qualitative examples that illustrate the previously observed trade-offs. The adversarial loss  degenerates for values $\lambda \geq 5$ and a stuttering phenomenon appears \cite{stuttering}. \autoref{tab:human_annotation} gathers results of human evaluation and show that our surrogates can better disentangle style while preserving more content than available methods.

\subsection{Adversarial Loss Fails to Disentangle  when $|\mathcal{Y}| \geq 3$}\label{ssec:msg}
In \autoref{fig:disent_sentiment_mst} we report the adversary accuracy of our different methods for the values of $\lambda$ using FYelp dataset with category label. In the binary setting for $\lambda \leq 1$, models using adversarial loss can learn disentangled representations while in the multi-class setting, the adversarial loss degenerates for small values of $\lambda$ (\textit{i.e} sentences are no longer fluent as shown by the increase in perplexity in \autoref{fig:ppl_category_st}). Minimizing MI based on our surrogates seems to mitigate the problem and offer a better control of the disentanglement degree for various values of $\lambda$ than $vCLUB-S$. Further results are gathered in \autoref{sec:addition_multu}.

\section{Summary and Concluding Remarks}
We devised a {new} alternative method to adversarial losses capable of learning  disentangled textual representation. Our method does not require  adversarial training and hence, it does not suffer in presence of multi-class setups. A key  feature of this method is to account for the  approximation error incurred when bounding the mutual information. Experiments show better trade-offs {than both adversarial training and \texttt{vCLUB-S}} on two fair classification tasks and demonstrate the efficiency to learn disentangled representations for sequence generation. As a matter of fact, there is no free-lunch for sentence generation tasks: \textit{although transferring style is easier with disentangled representations, it also removes important information about the content}. \\
The proposed method can replace the adversary in any kind of algorithms \cite{iclr_emnlp,style_transfert_1} with no modifications. Future work includes testing with other type of labels such as dialog act \cite{DBLP:conf/emnlp/ChapuisCMLC20,colombo2020guiding}, emotions \cite{DBLP:conf/wassa/WitonCMK18}, opinion \cite{garcia2019token} or speaker’s stance and confidence \cite{DBLP:conf/emnlp/DinkarCLC20}. Since it allows more fine-grained control over the amount of disentanglement, we expect it to be easier to tune when combined with more complex models.

\section{Acknowledgements} The authors would like to thanks Georg Pichler for the thorough reading. The work of Prof. Pablo Piantanida was supported by the European Commission's Marie Sklodowska-Curie Actions (MSCA), through the Marie Sklodowska-Curie IF (H2020-MSCAIF-2017-EF-797805).
\newpage

\bibliography{acl2021}

\begin{thebibliography}{74}
\expandafter\ifx\csname natexlab\endcsname\relax\def\natexlab#1{#1}\fi

\bibitem[{Ali and Silvey(1966)}]{kl}
Syed~Mumtaz Ali and Samuel~D Silvey. 1966.
\newblock A general class of coefficients of divergence of one distribution
  from another.
\newblock \emph{Journal of the Royal Statistical Society: Series B
  (Methodological)}, 28(1):131--142.

\bibitem[{Bao et~al.(2019)Bao, Zhou, Huang, Li, Mou, Vechtomova, Dai, and
  Chen}]{loss_1}
Yu~Bao, Hao Zhou, Shujian Huang, Lei Li, Lili Mou, Olga Vechtomova, Xinyu Dai,
  and Jiajun Chen. 2019.
\newblock Generating sentences from disentangled syntactic and semantic spaces.
\newblock \emph{arXiv preprint arXiv:1907.05789}.

\bibitem[{Barrett et~al.(2019)Barrett, Kementchedjhieva, Elazar, Elliott, and
  S{\o}gaard}]{adversarial_removal_2}
Maria Barrett, Yova Kementchedjhieva, Yanai Elazar, Desmond Elliott, and Anders
  S{\o}gaard. 2019.
\newblock Adversarial removal of demographic attributes revisited.
\newblock In \emph{Proceedings of the 2019 Conference on Empirical Methods in
  Natural Language Processing and the 9th International Joint Conference on
  Natural Language Processing (EMNLP-IJCNLP)}, pages 6331--6336.

\bibitem[{Belghazi et~al.(2018)Belghazi, Baratin, Rajeswar, Ozair, Bengio,
  Courville, and Hjelm}]{mine}
Mohamed~Ishmael Belghazi, Aristide Baratin, Sai Rajeswar, Sherjil Ozair, Yoshua
  Bengio, Aaron Courville, and R~Devon Hjelm. 2018.
\newblock Mine: mutual information neural estimation.
\newblock \emph{arXiv preprint arXiv:1801.04062}.

\bibitem[{{Bengio} et~al.(2013){Bengio}, {Courville}, and {Vincent}}]{6472238}
Y.~{Bengio}, A.~{Courville}, and P.~{Vincent}. 2013.
\newblock Representation learning: A review and new perspectives.
\newblock \emph{IEEE Transactions on Pattern Analysis and Machine
  Intelligence}, 35(8):1798--1828.

\bibitem[{Blodgett et~al.(2016)Blodgett, Green, and O{'}Connor}]{dial}
Su~Lin Blodgett, Lisa Green, and Brendan O{'}Connor. 2016.
\newblock \href {https://doi.org/10.18653/v1/D16-1120} {Demographic dialectal
  variation in social media: A case study of {A}frican-{A}merican {E}nglish}.
\newblock In \emph{Proceedings of the 2016 Conference on Empirical Methods in
  Natural Language Processing}, pages 1119--1130, Austin, Texas. Association
  for Computational Linguistics.

\bibitem[{Bojanowski et~al.(2017)Bojanowski, Grave, Joulin, and
  Mikolov}]{fasttext}
Piotr Bojanowski, Edouard Grave, Armand Joulin, and Tomas Mikolov. 2017.
\newblock Enriching word vectors with subword information.
\newblock \emph{Transactions of the Association for Computational Linguistics},
  5:135--146.

\bibitem[{Burgess et~al.(2018)Burgess, Higgins, Pal, Matthey, Watters,
  Desjardins, and Lerchner}]{conditional_generation}
Christopher~P Burgess, Irina Higgins, Arka Pal, Loic Matthey, Nick Watters,
  Guillaume Desjardins, and Alexander Lerchner. 2018.
\newblock Understanding disentangling in $\beta$-vae.
\newblock \emph{arXiv preprint arXiv:1804.03599}.

\bibitem[{Chapuis et~al.(2020)Chapuis, Colombo, Manica, Labeau, and
  Clavel}]{DBLP:conf/emnlp/ChapuisCMLC20}
Emile Chapuis, Pierre Colombo, Matteo Manica, Matthieu Labeau, and Chlo{\'{e}}
  Clavel. 2020.
\newblock \href {https://doi.org/10.18653/v1/2020.findings-emnlp.239}
  {Hierarchical pre-training for sequence labelling in spoken dialog}.
\newblock In \emph{Proceedings of the 2020 Conference on Empirical Methods in
  Natural Language Processing: Findings, {EMNLP} 2020, Online Event, 16-20
  November 2020}, pages 2636--2648. Association for Computational Linguistics.

\bibitem[{Cheng et~al.(2020{\natexlab{a}})Cheng, Hao, Dai, Liu, Gan, and
  Carin}]{cheng2020club}
Pengyu Cheng, Weituo Hao, Shuyang Dai, Jiachang Liu, Zhe Gan, and Lawrence
  Carin. 2020{\natexlab{a}}.
\newblock Club: A contrastive log-ratio upper bound of mutual information.
\newblock In \emph{International Conference on Machine Learning}, pages
  1779--1788. PMLR.

\bibitem[{Cheng et~al.(2020{\natexlab{b}})Cheng, Min, Shen, Malon, Zhang, Li,
  and Carin}]{dt_info}
Pengyu Cheng, Martin~Renqiang Min, Dinghan Shen, Christopher Malon, Yizhe
  Zhang, Yitong Li, and Lawrence Carin. 2020{\natexlab{b}}.
\newblock Improving disentangled text representation learning with
  information-theoretic guidance.
\newblock \emph{arXiv preprint arXiv:2006.00693}.

\bibitem[{Chung et~al.(2014)Chung, Gulcehre, Cho, and Bengio}]{gru}
Junyoung Chung, Caglar Gulcehre, KyungHyun Cho, and Yoshua Bengio. 2014.
\newblock Empirical evaluation of gated recurrent neural networks on sequence
  modeling.
\newblock \emph{arXiv preprint arXiv:1412.3555}.

\bibitem[{Colombo et~al.(2020)Colombo, Chapuis, Manica, Vignon, Varni, and
  Clavel}]{colombo2020guiding}
Pierre Colombo, Emile Chapuis, Matteo Manica, Emmanuel Vignon, Giovanna Varni,
  and Chloe Clavel. 2020.
\newblock Guiding attention in sequence-to-sequence models for dialogue act
  prediction.
\newblock In \emph{AAAI}, pages 7594--7601.

\bibitem[{Colombo et~al.(2019)Colombo, Witon, Modi, Kennedy, and
  Kapadia}]{colombo2019affect}
Pierre Colombo, Wojciech Witon, Ashutosh Modi, James Kennedy, and Mubbasir
  Kapadia. 2019.
\newblock Affect-driven dialog generation.
\newblock \emph{arXiv preprint arXiv:1904.02793}.

\bibitem[{Cover and Thomas(2006)}]{cover}
Thomas~M. Cover and Joy~A. Thomas. 2006.
\newblock \emph{Elements of Information Theory (Wiley Series in
  Telecommunications and Signal Processing)}.
\newblock Wiley-Interscience, USA.

\bibitem[{Daudel et~al.(2020)Daudel, Douc, and Portier}]{daudel:hal-02614605}
Kam{\'e}lia Daudel, Randal Douc, and Fran{\c c}ois Portier. 2020.
\newblock \href {https://hal.telecom-paris.fr/hal-02614605}
  {{Infinite-dimensional gradient-based descent for alpha-divergence
  minimisation}}.
\newblock Working paper or preprint.

\bibitem[{Denton et~al.(2017)}]{conditional_generation_1}
Emily~L Denton et~al. 2017.
\newblock Unsupervised learning of disentangled representations from video.
\newblock In \emph{Advances in neural information processing systems}, pages
  4414--4423.

\bibitem[{Dinkar et~al.(2020)Dinkar, Colombo, Labeau, and
  Clavel}]{DBLP:conf/emnlp/DinkarCLC20}
Tanvi Dinkar, Pierre Colombo, Matthieu Labeau, and Chlo{\'{e}} Clavel. 2020.
\newblock \href {https://doi.org/10.18653/v1/2020.emnlp-main.641} {The
  importance of fillers for text representations of speech transcripts}.
\newblock In \emph{Proceedings of the 2020 Conference on Empirical Methods in
  Natural Language Processing, {EMNLP} 2020, Online, November 16-20, 2020},
  pages 7985--7993. Association for Computational Linguistics.

\bibitem[{Donsker and Varadhan(1985)}]{donsker1985large}
MD~Donsker and SRS Varadhan. 1985.
\newblock Large deviations for stationary gaussian processes.
\newblock \emph{Communications in Mathematical Physics}, 97(1-2):187--210.

\bibitem[{Elazar and Goldberg(2018)}]{adversarial_removal}
Yanai Elazar and Yoav Goldberg. 2018.
\newblock Adversarial removal of demographic attributes from text data.
\newblock \emph{arXiv preprint arXiv:1808.06640}.

\bibitem[{Feutry et~al.(2018)Feutry, Piantanida, Bengio, and
  Duhamel}]{feutry2018learning}
Clément Feutry, Pablo Piantanida, Yoshua Bengio, and Pierre Duhamel. 2018.
\newblock \href {http://arxiv.org/abs/1802.09386} {Learning anonymized
  representations with adversarial neural networks}.

\bibitem[{Fu et~al.(2017)Fu, Tan, Peng, Zhao, and Yan}]{style_transfert_1}
Zhenxin Fu, Xiaoye Tan, Nanyun Peng, Dongyan Zhao, and Rui Yan. 2017.
\newblock Style transfer in text: Exploration and evaluation.
\newblock \emph{arXiv preprint arXiv:1711.06861}.

\bibitem[{Garcia et~al.(2019)Garcia, Colombo, Essid, d'Alch{\'e} Buc, and
  Clavel}]{garcia2019token}
Alexandre Garcia, Pierre Colombo, Slim Essid, Florence d'Alch{\'e} Buc, and
  Chlo{\'e} Clavel. 2019.
\newblock From the token to the review: A hierarchical multimodal approach to
  opinion mining.
\newblock \emph{arXiv preprint arXiv:1908.11216}.

\bibitem[{Hjelm et~al.(2018)Hjelm, Fedorov, Lavoie-Marchildon, Grewal, Bachman,
  Trischler, and Bengio}]{evo_mine}
R~Devon Hjelm, Alex Fedorov, Samuel Lavoie-Marchildon, Karan Grewal, Phil
  Bachman, Adam Trischler, and Yoshua Bengio. 2018.
\newblock Learning deep representations by mutual information estimation and
  maximization.
\newblock \emph{arXiv preprint arXiv:1808.06670}.

\bibitem[{Holtzman et~al.(2019)Holtzman, Buys, Du, Forbes, and
  Choi}]{stuttering}
Ari Holtzman, Jan Buys, Li~Du, Maxwell Forbes, and Yejin Choi. 2019.
\newblock The curious case of neural text degeneration.
\newblock \emph{arXiv preprint arXiv:1904.09751}.

\bibitem[{Hsieh et~al.(2018)Hsieh, Liu, Huang, Fei-Fei, and Niebles}]{video}
Jun-Ting Hsieh, Bingbin Liu, De-An Huang, Li~F Fei-Fei, and Juan~Carlos
  Niebles. 2018.
\newblock Learning to decompose and disentangle representations for video
  prediction.
\newblock In \emph{Advances in Neural Information Processing Systems}, pages
  517--526.

\bibitem[{Hu et~al.(2017)Hu, Yang, Liang, Salakhutdinov, and Xing}]{loss_5}
Zhiting Hu, Zichao Yang, Xiaodan Liang, Ruslan Salakhutdinov, and Eric~P Xing.
  2017.
\newblock Toward controlled generation of text.
\newblock \emph{arXiv preprint arXiv:1703.00955}.

\bibitem[{Hung et~al.(2018)Hung, Chen, and Yang}]{audio}
Yun-Ning Hung, Yi-An Chen, and Yi-Hsuan Yang. 2018.
\newblock Learning disentangled representations for timber and pitch in music
  audio.
\newblock \emph{arXiv preprint arXiv:1811.03271}.

\bibitem[{Jain et~al.(2019)Jain, Mishra, Azad, and Sankaranarayanan}]{loss_3}
Parag Jain, Abhijit Mishra, Amar~Prakash Azad, and Karthik Sankaranarayanan.
  2019.
\newblock Unsupervised controllable text formalization.
\newblock In \emph{Proceedings of the AAAI Conference on Artificial
  Intelligence}, volume~33, pages 6554--6561.

\bibitem[{Jalalzai et~al.(2020)Jalalzai, Colombo, Clavel, Gaussier, Varni,
  Vignon, and Sabourin}]{jalalzai2020heavy}
Hamid Jalalzai, Pierre Colombo, Chlo{\'e} Clavel, Eric Gaussier, Giovanna
  Varni, Emmanuel Vignon, and Anne Sabourin. 2020.
\newblock Heavy-tailed representations, text polarity classification \& data
  augmentation.
\newblock \emph{arXiv preprint arXiv:2003.11593}.

\bibitem[{John et~al.(2018)John, Mou, Bahuleyan, and Vechtomova}]{text}
Vineet John, Lili Mou, Hareesh Bahuleyan, and Olga Vechtomova. 2018.
\newblock Disentangled representation learning for non-parallel text style
  transfer.
\newblock \emph{arXiv preprint arXiv:1808.04339}.

\bibitem[{Joulin et~al.(2016{\natexlab{a}})Joulin, Grave, Bojanowski, Douze,
  J{\'e}gou, and Mikolov}]{fastext_2}
Armand Joulin, Edouard Grave, Piotr Bojanowski, Matthijs Douze, H{\'e}rve
  J{\'e}gou, and Tomas Mikolov. 2016{\natexlab{a}}.
\newblock Fasttext.zip: Compressing text classification models.
\newblock \emph{arXiv preprint arXiv:1612.03651}.

\bibitem[{Joulin et~al.(2016{\natexlab{b}})Joulin, Grave, Bojanowski, and
  Mikolov}]{fastext_1}
Armand Joulin, Edouard Grave, Piotr Bojanowski, and Tomas Mikolov.
  2016{\natexlab{b}}.
\newblock Bag of tricks for efficient text classification.
\newblock \emph{arXiv preprint arXiv:1607.01759}.

\bibitem[{Kingma and Ba(2014)}]{adam}
Diederik~P Kingma and Jimmy Ba. 2014.
\newblock Adam: A method for stochastic optimization.
\newblock \emph{arXiv preprint arXiv:1412.6980}.

\bibitem[{Kinney and Atwal(2014)}]{equitable}
Justin~B Kinney and Gurinder~S Atwal. 2014.
\newblock Equitability, mutual information, and the maximal information
  coefficient.
\newblock \emph{Proceedings of the National Academy of Sciences},
  111(9):3354--3359.

\bibitem[{Kpotufe(2017)}]{pmlr-v54-kpotufe17a}
Samory Kpotufe. 2017.
\newblock \href {http://proceedings.mlr.press/v54/kpotufe17a.html} {{Lipschitz
  Density-Ratios, Structured Data, and Data-driven Tuning}}.
\newblock volume~54 of \emph{Proceedings of Machine Learning Research}, pages
  1320--1328, Fort Lauderdale, FL, USA. PMLR.

\bibitem[{Krippendorff(2018)}]{krippendorff2018content}
Klaus Krippendorff. 2018.
\newblock \emph{Content analysis: An introduction to its methodology}.
\newblock Sage publications.

\bibitem[{Kudo(2018)}]{tok_0}
Taku Kudo. 2018.
\newblock Subword regularization: Improving neural network translation models
  with multiple subword candidates.
\newblock \emph{arXiv preprint arXiv:1804.10959}.

\bibitem[{Kumar~Verma et~al.(2018)Kumar~Verma, Arora, Mishra, and
  Rai}]{few_shot}
Vinay Kumar~Verma, Gundeep Arora, Ashish Mishra, and Piyush Rai. 2018.
\newblock Generalized zero-shot learning via synthesized examples.
\newblock In \emph{Proceedings of the IEEE conference on computer vision and
  pattern recognition}, pages 4281--4289.

\bibitem[{Lample et~al.(2018)Lample, Subramanian, Smith, Denoyer, Ranzato, and
  Boureau}]{multiple}
Guillaume Lample, Sandeep Subramanian, Eric Smith, Ludovic Denoyer,
  Marc'Aurelio Ranzato, and Y-Lan Boureau. 2018.
\newblock Multiple-attribute text rewriting.
\newblock In \emph{International Conference on Learning Representations}.

\bibitem[{Li et~al.(2015)Li, Galley, Brockett, Gao, and Dolan}]{mmi}
Jiwei Li, Michel Galley, Chris Brockett, Jianfeng Gao, and Bill Dolan. 2015.
\newblock A diversity-promoting objective function for neural conversation
  models.
\newblock \emph{arXiv preprint arXiv:1510.03055}.

\bibitem[{Li et~al.(2018)Li, Jia, He, and Liang}]{li2018delete}
Juncen Li, Robin Jia, He~He, and Percy Liang. 2018.
\newblock Delete, retrieve, generate: A simple approach to sentiment and style
  transfer.
\newblock \emph{arXiv preprint arXiv:1804.06437}.

\bibitem[{Loshchilov and Hutter(2017)}]{adamw}
Ilya Loshchilov and Frank Hutter. 2017.
\newblock Decoupled weight decay regularization.
\newblock \emph{arXiv preprint arXiv:1711.05101}.

\bibitem[{Mohri et~al.(2019)Mohri, Sivek, and Suresh}]{fair_3}
Mehryar Mohri, Gary Sivek, and Ananda~Theertha Suresh. 2019.
\newblock Agnostic federated learning.
\newblock \emph{arXiv preprint arXiv:1902.00146}.

\bibitem[{Oord et~al.(2018)Oord, Li, and Vinyals}]{nce}
Aaron van~den Oord, Yazhe Li, and Oriol Vinyals. 2018.
\newblock Representation learning with contrastive predictive coding.
\newblock \emph{arXiv preprint arXiv:1807.03748}.

\bibitem[{Paninski(2003)}]{estimation}
Liam Paninski. 2003.
\newblock Estimation of entropy and mutual information.
\newblock \emph{Neural computation}, 15(6):1191--1253.

\bibitem[{Pichler et~al.(2020)Pichler, Piantanida, and
  Koliander}]{pichler2020estimation}
Georg Pichler, Pablo Piantanida, and Günther Koliander. 2020.
\newblock \href {http://arxiv.org/abs/2002.02851} {On the estimation of
  information measures of continuous distributions}.

\bibitem[{Post(2018)}]{sacrebleu}
Matt Post. 2018.
\newblock A call for clarity in reporting bleu scores.
\newblock \emph{arXiv preprint arXiv:1804.08771}.

\bibitem[{Prabhumoye et~al.(2018)Prabhumoye, Tsvetkov, Salakhutdinov, and
  Black}]{back}
Shrimai Prabhumoye, Yulia Tsvetkov, Ruslan Salakhutdinov, and Alan~W Black.
  2018.
\newblock Style transfer through back-translation.
\newblock \emph{arXiv preprint arXiv:1804.09000}.

\bibitem[{Radford et~al.(2019)Radford, Wu, Child, Luan, Amodei, and
  Sutskever}]{gpt}
Alec Radford, Jeffrey Wu, Rewon Child, David Luan, Dario Amodei, and Ilya
  Sutskever. 2019.
\newblock Language models are unsupervised multitask learners.
\newblock \emph{OpenAI Blog}, 1(8):9.

\bibitem[{Raffel et~al.(2019)Raffel, Shazeer, Roberts, Lee, Narang, Matena,
  Zhou, Li, and Liu}]{t2t}
Colin Raffel, Noam Shazeer, Adam Roberts, Katherine Lee, Sharan Narang, Michael
  Matena, Yanqi Zhou, Wei Li, and Peter~J Liu. 2019.
\newblock Exploring the limits of transfer learning with a unified text-to-text
  transformer.
\newblock \emph{arXiv preprint arXiv:1910.10683}.

\bibitem[{R{\'e}nyi et~al.(1961)}]{renyi1961measures}
Alfr{\'e}d R{\'e}nyi et~al. 1961.
\newblock On measures of entropy and information.
\newblock In \emph{Proceedings of the Fourth Berkeley Symposium on Mathematical
  Statistics and Probability, Volume 1: Contributions to the Theory of
  Statistics}. The Regents of the University of California.

\bibitem[{Rosenblatt(1958)}]{perceptron}
Frank Rosenblatt. 1958.
\newblock The perceptron: a probabilistic model for information storage and
  organization in the brain.
\newblock \emph{Psychological review}, 65(6):386.

\bibitem[{Sanchez et~al.(2019)Sanchez, Serrurier, and Ortner}]{iclr_irrelevant}
Eduardo~Hugo Sanchez, Mathieu Serrurier, and Mathias Ortner. 2019.
\newblock Learning disentangled representations via mutual information
  estimation.
\newblock \emph{arXiv preprint arXiv:1912.03915}.

\bibitem[{Sennrich et~al.(2016)Sennrich, Haddow, and Birch}]{tok_1}
Rico Sennrich, Barry Haddow, and Alexandra Birch. 2016.
\newblock \href {https://doi.org/10.18653/v1/P16-1162} {Neural machine
  translation of rare words with subword units}.
\newblock In \emph{Proceedings of the 54th Annual Meeting of the Association
  for Computational Linguistics (Volume 1: Long Papers)}, pages 1715--1725,
  Berlin, Germany. Association for Computational Linguistics.

\bibitem[{Srivastava et~al.(2014)Srivastava, Hinton, Krizhevsky, Sutskever, and
  Salakhutdinov}]{dropout}
Nitish Srivastava, Geoffrey Hinton, Alex Krizhevsky, Ilya Sutskever, and Ruslan
  Salakhutdinov. 2014.
\newblock Dropout: a simple way to prevent neural networks from overfitting.
\newblock \emph{The journal of machine learning research}, 15(1):1929--1958.

\bibitem[{van Steenkiste et~al.(2019)van Steenkiste, Locatello, Schmidhuber,
  and Bachem}]{visual}
Sjoerd van Steenkiste, Francesco Locatello, J{\"u}rgen Schmidhuber, and Olivier
  Bachem. 2019.
\newblock Are disentangled representations helpful for abstract visual
  reasoning?
\newblock In \emph{Advances in Neural Information Processing Systems}, pages
  14245--14258.

\bibitem[{Sugiyama et~al.(2012)Sugiyama, Suzuki, and
  Kanamori}]{10.5555/2181148}
Masashi Sugiyama, Taiji Suzuki, and Takafumi Kanamori. 2012.
\newblock \emph{Density Ratio Estimation in Machine Learning}, 1st edition.
\newblock Cambridge University Press, USA.

\bibitem[{Sutskever et~al.(2014)Sutskever, Vinyals, and Le}]{seq2seq}
Ilya Sutskever, Oriol Vinyals, and Quoc~V Le. 2014.
\newblock Sequence to sequence learning with neural networks.
\newblock In \emph{Advances in neural information processing systems}, pages
  3104--3112.

\bibitem[{Tikhonov et~al.(2019)Tikhonov, Shibaev, Nagaev, Nugmanova, and
  Yamshchikov}]{iclr_emnlp}
Alexey Tikhonov, Viacheslav Shibaev, Aleksander Nagaev, Aigul Nugmanova, and
  Ivan~P Yamshchikov. 2019.
\newblock Style transfer for texts: Retrain, report errors, compare with
  rewrites.
\newblock In \emph{Proceedings of the 2019 Conference on Empirical Methods in
  Natural Language Processing and the 9th International Joint Conference on
  Natural Language Processing (EMNLP-IJCNLP)}, pages 3927--3936.

\bibitem[{Van~Erven and Harremos(2014)}]{survey}
Tim Van~Erven and Peter Harremos. 2014.
\newblock R{\'e}nyi divergence and kullback-leibler divergence.
\newblock \emph{IEEE Transactions on Information Theory}, 60(7):3797--3820.

\bibitem[{Wieting et~al.(2017)Wieting, Mallinson, and
  Gimpel}]{back_translation_paper}
John Wieting, Jonathan Mallinson, and Kevin Gimpel. 2017.
\newblock Learning paraphrastic sentence embeddings from back-translated
  bitext.
\newblock \emph{arXiv preprint arXiv:1706.01847}.

\bibitem[{Witon et~al.(2018)Witon, Colombo, Modi, and
  Kapadia}]{DBLP:conf/wassa/WitonCMK18}
Wojciech Witon, Pierre Colombo, Ashutosh Modi, and Mubbasir Kapadia. 2018.
\newblock \href {https://doi.org/10.18653/v1/w18-6236} {Disney at {IEST} 2018:
  Predicting emotions using an ensemble}.
\newblock In \emph{Proceedings of the 9th Workshop on Computational Approaches
  to Subjectivity, Sentiment and Social Media Analysis, WASSA@EMNLP 2018,
  Brussels, Belgium, October 31, 2018}, pages 248--253. Association for
  Computational Linguistics.

\bibitem[{Wolf et~al.(2019)Wolf, Debut, Sanh, Chaumond, Delangue, Moi, Cistac,
  Rault, Louf, Funtowicz, Davison, Shleifer, von Platen, Ma, Jernite, Plu, Xu,
  Scao, Gugger, Drame, Lhoest, and Rush}]{hf}
Thomas Wolf, Lysandre Debut, Victor Sanh, Julien Chaumond, Clement Delangue,
  Anthony Moi, Pierric Cistac, Tim Rault, Rémi Louf, Morgan Funtowicz, Joe
  Davison, Sam Shleifer, Patrick von Platen, Clara Ma, Yacine Jernite, Julien
  Plu, Canwen Xu, Teven~Le Scao, Sylvain Gugger, Mariama Drame, Quentin Lhoest,
  and Alexander~M. Rush. 2019.
\newblock Huggingface's transformers: State-of-the-art natural language
  processing.
\newblock \emph{ArXiv}, abs/1910.03771.

\bibitem[{Xie et~al.(2017)Xie, Dai, Du, Hovy, and Neubig}]{adv_classif_fair_1}
Qizhe Xie, Zihang Dai, Yulun Du, Eduard Hovy, and Graham Neubig. 2017.
\newblock Controllable invariance through adversarial feature learning.
\newblock In \emph{Advances in Neural Information Processing Systems}, pages
  585--596.

\bibitem[{Xu et~al.(2015)Xu, Wang, Chen, and Li}]{leaky}
Bing Xu, Naiyan Wang, Tianqi Chen, and Mu~Li. 2015.
\newblock Empirical evaluation of rectified activations in convolutional
  network.
\newblock \emph{arXiv preprint arXiv:1505.00853}.

\bibitem[{Xu et~al.(2019)Xu, Ge, and Wei}]{formality}
Ruochen Xu, Tao Ge, and Furu Wei. 2019.
\newblock Formality style transfer with hybrid textual annotations.
\newblock \emph{arXiv preprint arXiv:1903.06353}.

\bibitem[{Yamshchikov et~al.(2019)Yamshchikov, Shibaev, Nagaev, Jost, and
  Tikhonov}]{iclr_workshop}
Ivan~P Yamshchikov, Viacheslav Shibaev, Aleksander Nagaev, J{\"u}rgen Jost, and
  Alexey Tikhonov. 2019.
\newblock Decomposing textual information for style transfer.
\newblock \emph{arXiv preprint arXiv:1909.12928}.

\bibitem[{Yi et~al.(2020)Yi, Liu, Li, and Sun}]{loss_2}
Xiaoyuan Yi, Zhenghao Liu, Wenhao Li, and Maosong Sun. 2020.
\newblock \href {https://doi.org/10.24963/ijcai.2020/526} {Text style transfer
  via learning style instance supported latent space}.
\newblock In \emph{Proceedings of the Twenty-Ninth International Joint
  Conference on Artificial Intelligence, {IJCAI-20}}, pages 3801--3807.
  International Joint Conferences on Artificial Intelligence Organization.

\bibitem[{Zafar et~al.(2017)Zafar, Valera, Gomez~Rodriguez, and
  Gummadi}]{fair_2}
Muhammad~Bilal Zafar, Isabel Valera, Manuel Gomez~Rodriguez, and Krishna~P
  Gummadi. 2017.
\newblock Fairness beyond disparate treatment \& disparate impact: Learning
  classification without disparate mistreatment.
\newblock In \emph{Proceedings of the 26th international conference on world
  wide web}, pages 1171--1180.

\bibitem[{Zemel et~al.(2013)Zemel, Wu, Swersky, Pitassi, and Dwork}]{fair_1}
Rich Zemel, Yu~Wu, Kevin Swersky, Toni Pitassi, and Cynthia Dwork. 2013.
\newblock \href {http://proceedings.mlr.press/v28/zemel13.html} {Learning fair
  representations}.
\newblock volume~28 of \emph{Proceedings of Machine Learning Research}, pages
  325--333, Atlanta, Georgia, USA. PMLR.

\bibitem[{Zhang et~al.(2018)Zhang, Ding, and Soricut}]{loss_4}
Ye~Zhang, Nan Ding, and Radu Soricut. 2018.
\newblock Shaped: Shared-private encoder-decoder for text style adaptation.
\newblock \emph{arXiv preprint arXiv:1804.04093}.

\bibitem[{Zhang et~al.(2020)Zhang, Ge, and Sun}]{formality_1}
Yi~Zhang, Tao Ge, and Xu~Sun. 2020.
\newblock Parallel data augmentation for formality style transfer.
\newblock \emph{arXiv preprint arXiv:2005.07522}.

\bibitem[{Zhu et~al.(2015)Zhu, Kiros, Zemel, Salakhutdinov, Urtasun, Torralba,
  and Fidler}]{bookcorpus}
Yukun Zhu, Ryan Kiros, Rich Zemel, Ruslan Salakhutdinov, Raquel Urtasun,
  Antonio Torralba, and Sanja Fidler. 2015.
\newblock Aligning books and movies: Towards story-like visual explanations by
  watching movies and reading books.
\newblock In \emph{Proceedings of the IEEE international conference on computer
  vision}, pages 19--27.

\end{thebibliography}
\bibliographystyle{acl_natbib}

\newpage
\appendix
\section{Additional Details On the Surrogates}\label{sec:proofs}

\subsection{Proof of Inequality  \autoref{eq:adv_losses_mi}}\label{sec:mi}
In this section, we provide a formal proof of the \autoref{eq:adv_losses_mi}.  Let $(Z, Y)$ be an arbitrary pair of RVs with $(Z, Y) \sim p_{ZY}$ according to some underlying  pdf, and let $q_{\widehat{Y}|Z}$ be a conditional variational probability distribution on the discrete attributes satisfying $p_{ZY} \ll p_Z\cdot  q_{\widehat{Y}|Z}$, i.e., absolutely continuous.
\begin{equation}
    I(Z;Y) \geq H(Y) - \textrm{CE}(\widehat{Y}|Z).
\end{equation}

\textit{Proof:} We start by the definition of the MI and use the fact that the maximum entropy distribution is reached for the uniform law in the case of a discrete variable (see \cite{cover}).
\begin{align}
        I(Z;Y) &= H(Y) - H(Y|Z) \\
       & = \textrm{Const}  - H(Y|Z).
\end{align}

We then need to find the relationship between the cross-entropy and the conditional entropy. 

\begin{align}\label{eq:bound_Ce_ce} 
    \begin{split}
   & \textrm{KL}(p_{YZ}\|q_{\widehat{Y}Z})  \\
   &= E_{YZ}\left[\log \frac{p_{Y|Z}(Y|Z)}{q_{\widehat{Y}|Z}(Y|Z)}\right] \\
    &= E_{YZ}\big[\log p_{{Y}|Z}(Y|Z) \big] - E_{YZ}\big[\log q_{\widehat{Y}|Z}(Y|Z) \big] \\
     &= - H(Y|Z) + \textrm{CE}(\widehat{Y}|Z).  
         \end{split}
\end{align}
We know that $ \textrm{KL}(p_{YZ}\|q_{\widehat{Y}Z}) \geq 0$, thus $\textrm{CE}(\widehat{Y}|Z)  \geq H(Y|Z)$ which gives the result.

The underlying hypothesis made by approximating the MI with an adversarial loss is that the contribution of gradient from $\textrm{KL}(p_{YZ}\|q_{\widehat{Y}Z})$ to the bound is negligible.

\subsection{Proof of \autoref{th:one}}
Let $(Z, Y)$ be an arbitrary pair of RVs with $(Z, Y) \sim p_{ZY}$ according to some underlying  pdf, and let $q_{\widehat{Y}|Z}$ be a conditional variational probability distribution satisfying $p_{ZY} \ll p_Z\cdot  q_{\widehat{Y}|Z}$, i.e., absolutely continuous. To obtain an upper bound on the MI we need to upper bound the entropy $H(Y)$ and to lower bound the conditional entropy $H(Y|Z)$. 

\textbf{Upper bound on $H(Y)$}. Since the KL divergence is non-negative, we have 

\begin{align}
    &H(Y)  \leq \mathbb{E}_Y \left[-\log q_Y(Y) \right]  \\
    & = \mathbb{E}_Y \left[-\log \int q_{\widehat{Y}|Z}(Y|z) p_Z (z)dz \right]. 
\end{align}

\textbf{Lower bounds on $H(Y|Z)$.} We have the following inequalities:
\begin{align}\label{eq:kl_bound}
\begin{split}
    H(Y|Z) &= \mathbb{E}_{YZ} \left[-\log q_{\widehat{Y}|Z}(Y|Z)\right] - \\& \textrm{KL}(p_{YZ}\| p_Z \cdot q_{\widehat{Y}|Z}),
\end{split}
\end{align}
where $\textrm{KL}(p_{YZ}\| p_Z \cdot q_{\widehat{Y}|Z})$ denotes the KL divergence. Furthermore, for arbitrary values $\alpha > 1$,
\begin{align}
\begin{split}\label{eq:renyi_bound}
    H(Y|Z) \leq &\mathbb{E}_{YZ} \left[-\log q_{\widehat{Y}|Z}(Y|Z)\right] - \\ & D_\alpha(p_{YZ}\| p_Z \cdot q_{\widehat{Y}|Z}),
\end{split}
\end{align}
where ${D}_\alpha(p_{YZ}\| p_Z \cdot q_{\widehat{Y}|Z}) =$
$$
 \frac{1}{\alpha - 1} \log \mathbb{E}_{ZY}\left[R^{\alpha-1}(Z,Y)\right]
$$
is the Renyi divergence with
$$
R(y, z) = \frac{p_{Y|Z}(y|z)}{q_{\widehat{Y}|Z}(y|z)}. 
$$
The proof of \autoref{eq:kl_bound} is given in  \autoref{sec:mi}. In order to show \autoref{eq:renyi_bound}, we remark that Renyi divergence is non-decreasing function $ \alpha \mapsto {D}_\alpha(p_{ZY}\|p_Z \cdot  q_{\widehat{Y}|Z})$  in $\alpha \in [0,+\infty)$ (the reader is refereed to \cite{survey} for a detailed proof). Thus, we have $\forall \alpha >1$, 
\begin{equation}
\textrm{KL}(p_{ZY}\|p_Z \cdot q_{\widehat{Y}|Z}) \leq  {D}_\alpha(p_{ZY}\|p_Z\cdot  q_{\widehat{Y}|Z}). 
\end{equation} 
Therefore, from expression \autoref{eq:kl_bound} we obtain the desired result.

\subsection{Optimization of the Surrogates on MI}\label{sec:practical_surrogate}

In this section, we give details to facilitate the practical implementation of our methods.

\subsubsection{Computing the entropy $H(Y)$}
\begin{equation}
\begin{aligned}
 H(Y)  &  \leq  \mathbb{E}_Y \left[-\log \int
 q_{\widehat{Y}|Z}(Y|z) p_Z (z)dz  \right] \\ & \approx \mathbb{E}_Y \left[-\log \sum_{i=1}^n q_{\widehat{Y}|Z}(Y|z_i) \right] + \textrm{const.} \\
    & \approx - \frac{1}{|\mathcal{Y}|} \sum_{j=1}^{|\mathcal{Y}|}  \log  \sum_{i=1}^n C_{\theta_c}(z_i)_{y_j}  + \textrm{const.}
\end{aligned}    
\end{equation}
where $C_{\theta_c}(z_i)_{y_j}$ is the $y_j$-th component of the normalised output of the classifier $C_{\theta_c}$.  

\subsubsection{Computing the lower bound on $H(Y|Z)$}
The upper bound helds for $\alpha > 1$, 
    \begin{align}
    \begin{split}
        H(Y|Z)
        & \approx  \textrm{CE}(Y|Z) - \widehat{D}_\alpha(p_{ZY}\| p_Z \cdot  q_{\widehat{Y}|Z}) \\
         & \approx  -\frac{1}{n} \sum_{i=1}^n \log q_{\widehat{Y}|Z}(y_i|z_i) - \\& \frac{1}{\alpha - 1} \log  \sum_{i=1}^n R^{\alpha-1}(z_i,y_i).
           \end{split}
    \end{align}

\textbf{Estimating the density-ratio $R(z, y)$}
In what follows we apply the so-called density-ratio trick to our specific setup. Suppose we have a balanced dataset $\{(y^p_i,z^p_i)\} \sim p_{YZ}$ and $\{(y^q_i,z^q_i)\} \sim q_{\widehat{Y}|Z}p_Z$ with $i \in [1,K]$. The density-ratio trick consists in training a classifier $C_{\theta_R}$ to distinguish between theses two distribution. Samples coming from $p$ are labelled $u=1$, samples coming from $q$ are labelled $u=0$. 
Thus, we can rewrite $R(z,y)$ as 
    \begin{align}
  R(z, y) &= \frac{p_{Y|Z}(y,z)}{q_{\widehat{Y}|Z}(y,z)}  \\
   &= \frac{p_{YZ|U}(y,z|u=0)}{p_{YZ|U}(y,z|u=1)}  \\
    &= \frac{p_{U|YZ}(u=0|y,z)}{p_{U|YZ}(u=1|y,z)}  \frac{p_U(u= 1)}{p_U(u= 0)}  \\
  & =\frac{p_{U|YZ}(u=0|y,z)}{p_{U|YZ}(u=1|y,z)} \\
  & =\frac{p_{U|YZ}(u=0|y,z)}{1 - p_{U|YZ}(u=0|y,z)}. 
     \end{align}
Obviously, the true posterior distribution $p_{U|YZ}$ is unknown. However, if $C_{\theta_R}$ is well trained, then $p_{U|YZ}(u=0|y,z) \approx \sigma(C_{\theta_R}(y,z))$, where $\sigma(\cdot)$ denotes the sigmoid function.
A detailled procedure for training is given in Algorithm~\ref{alg:optimization}.

\begin{algorithm}[t]
\caption{Our method for the fair classification task}
\begin{algorithmic}[1]
\INPUT  training dataset for the encoder $\mathcal{D}_n=\{(x_1,y_1,l_1),\; \ldots,\; (x_n,y_n,l_n)\}$, batch size $m$,  training dataset for the classifiers and decoder $\mathcal{D^\prime}_n=\{(x^\prime_1,y^\prime_1,l^\prime_1),\; \ldots,\; (x^\prime_n,y^\prime_n,l^\prime_n)\}$. 
\Initialize parameters $(\theta_e, \theta_R,\theta_c, \theta_d)$ of the encoder $f_{\theta_e}$, classifiers $C_{\theta_R}$, $C_{\theta_c}$, $f_{\theta_d}$
\Optimization
\While{$(\theta_e, \theta_R,\theta_c, \theta_d)$ not converged}
\For{\texttt{ $i \in [1,Unroll]$}} \Comment{Train $C_{\theta_c}$, $C_{\theta_R}$, $f_{\theta_d}$}
        \State Sample a batch $\mathcal{B}^\prime$ from $\mathcal{D^\prime}$
        \State Update ${\theta_R}$ based  $\mathcal{B}^\prime$  and using $C_{\theta_c}$
        \State Update ${\theta_c}$ with  $\mathcal{B}^\prime$ 
        \State Update ${\theta_d}$ with  $\mathcal{B}^\prime$ 
      \EndFor
\State Sample a batch $\mathcal{B}$ from $\mathcal{D}$ \Comment{Train $f_{\theta_e}$}
\State Update $\theta_e$ with $\mathcal{B}$ using \autoref{eq:all_loss} with ${\theta_d}$.
\EndWhile
\OUTPUT $f_{\theta_e}$, $f_{\theta_d}$
\end{algorithmic}
\label{alg:optimization}
\end{algorithm}

\begin{figure*}
\centering     
\begin{subfigure}[t]{0.3\textwidth}
        \includegraphics[width=\textwidth]{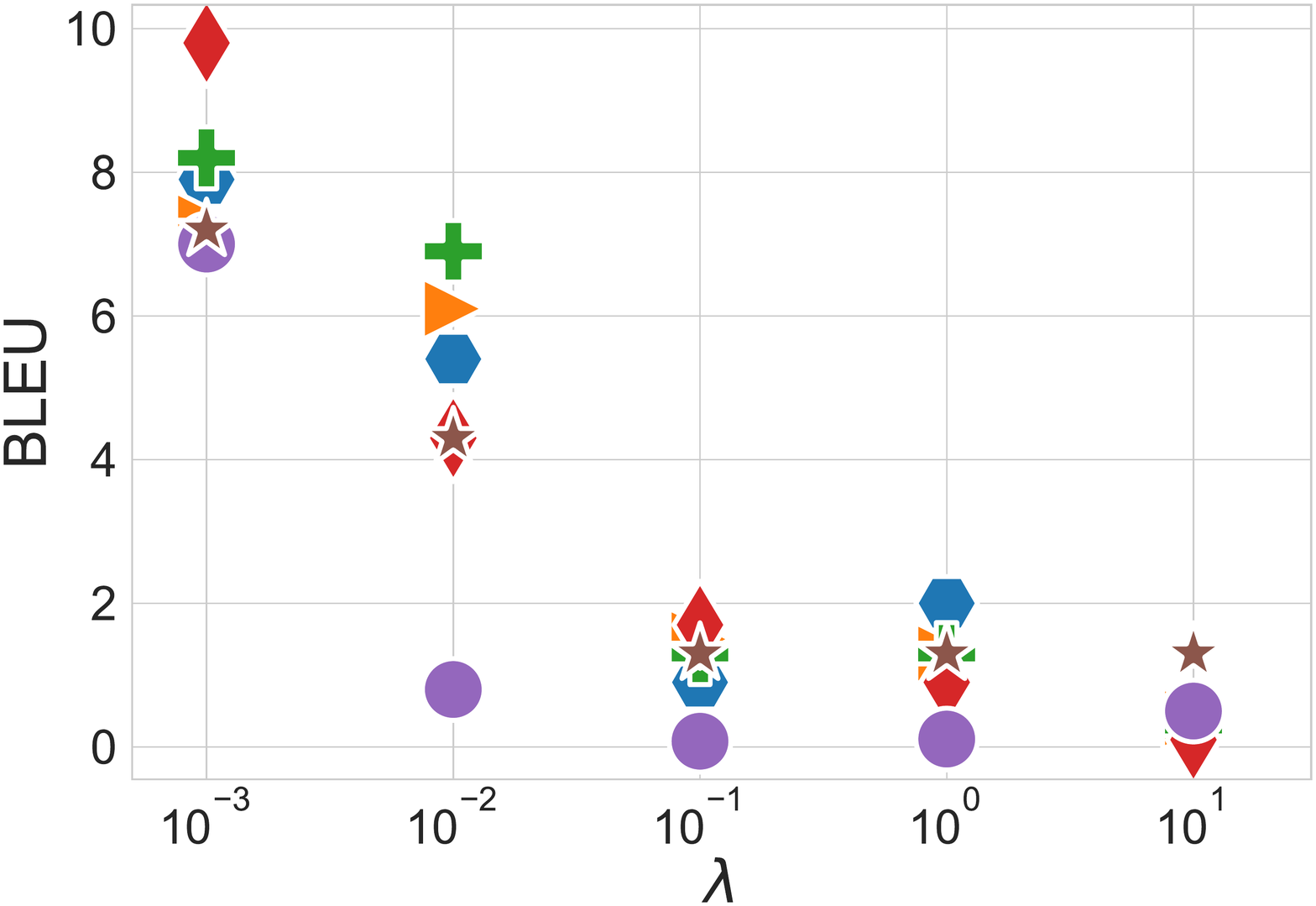}
        \caption{}\label{fig:bleu_category_st} 
    \end{subfigure} 
    \begin{subfigure}[t]{0.3\textwidth}
        \includegraphics[width=\textwidth]{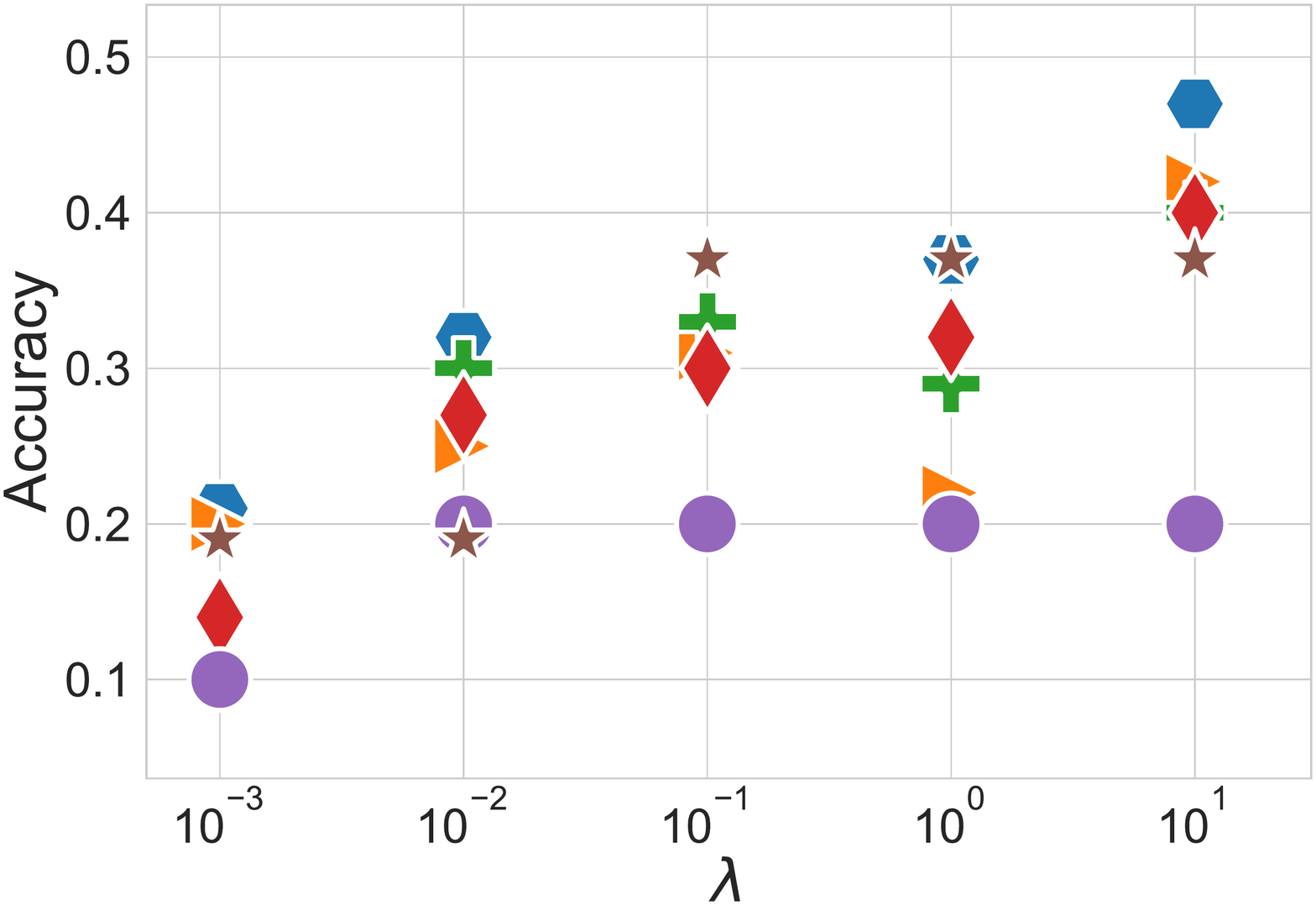}
        \caption{}\label{fig:accuracy_category_st} 
    \end{subfigure} 
    \begin{subfigure}[t]{0.3\textwidth}
        \includegraphics[width=\textwidth]{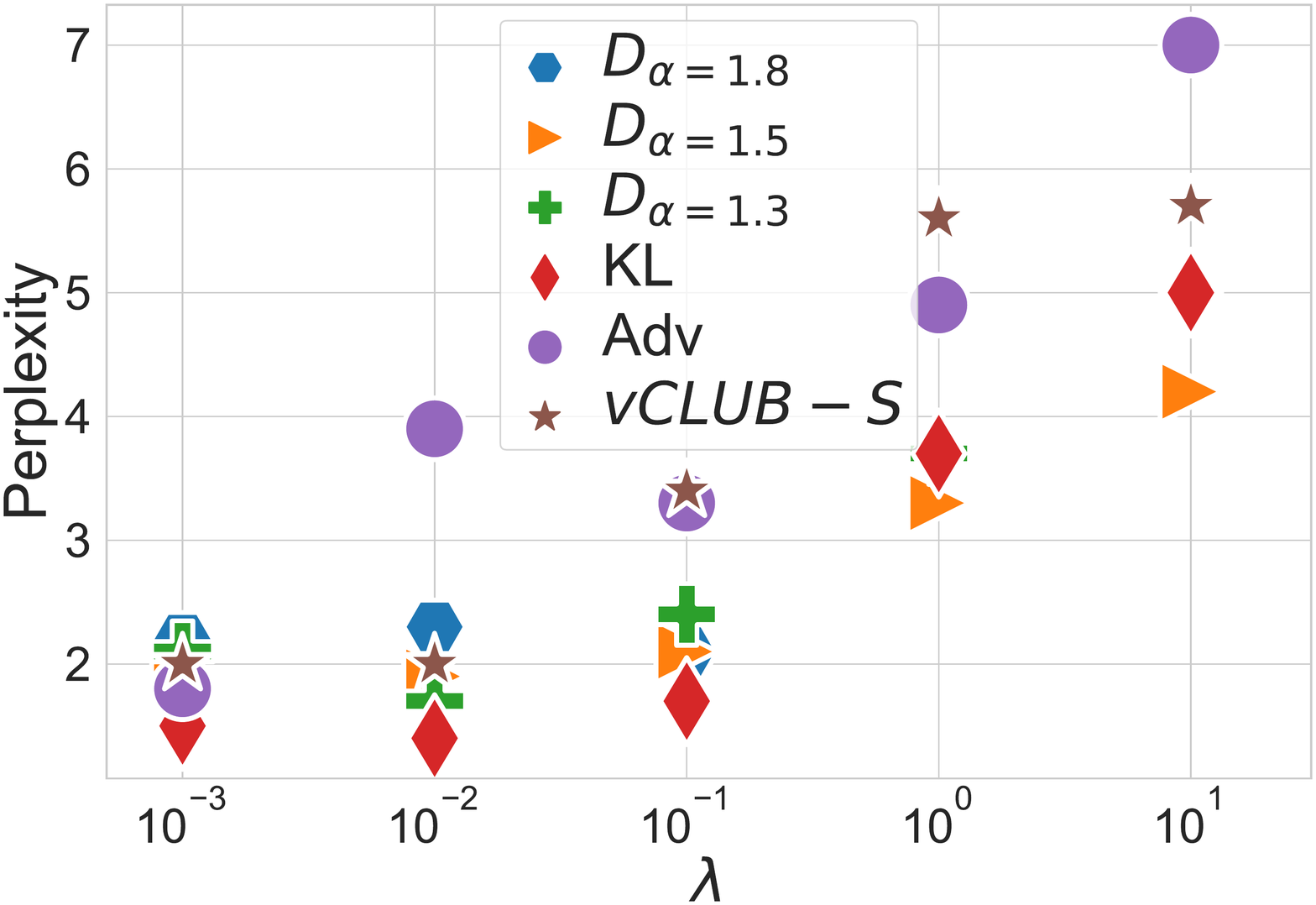}
         \caption{}\label{fig:ppl_category_st}
    \end{subfigure}
            \caption{Numerical experiments on multiclass style transfer using categorical labels. Results include: BLEU (\autoref{fig:bleu_category_st})); style transfer accuracy (\autoref{fig:accuracy_category_st}); sentence fluency (\autoref{fig:ppl_category_st}).}\label{fig:style_category}
    
\end{figure*}
\begin{figure}
    \begin{subfigure}[t]{0.4\textwidth}
        \includegraphics[width=\textwidth]{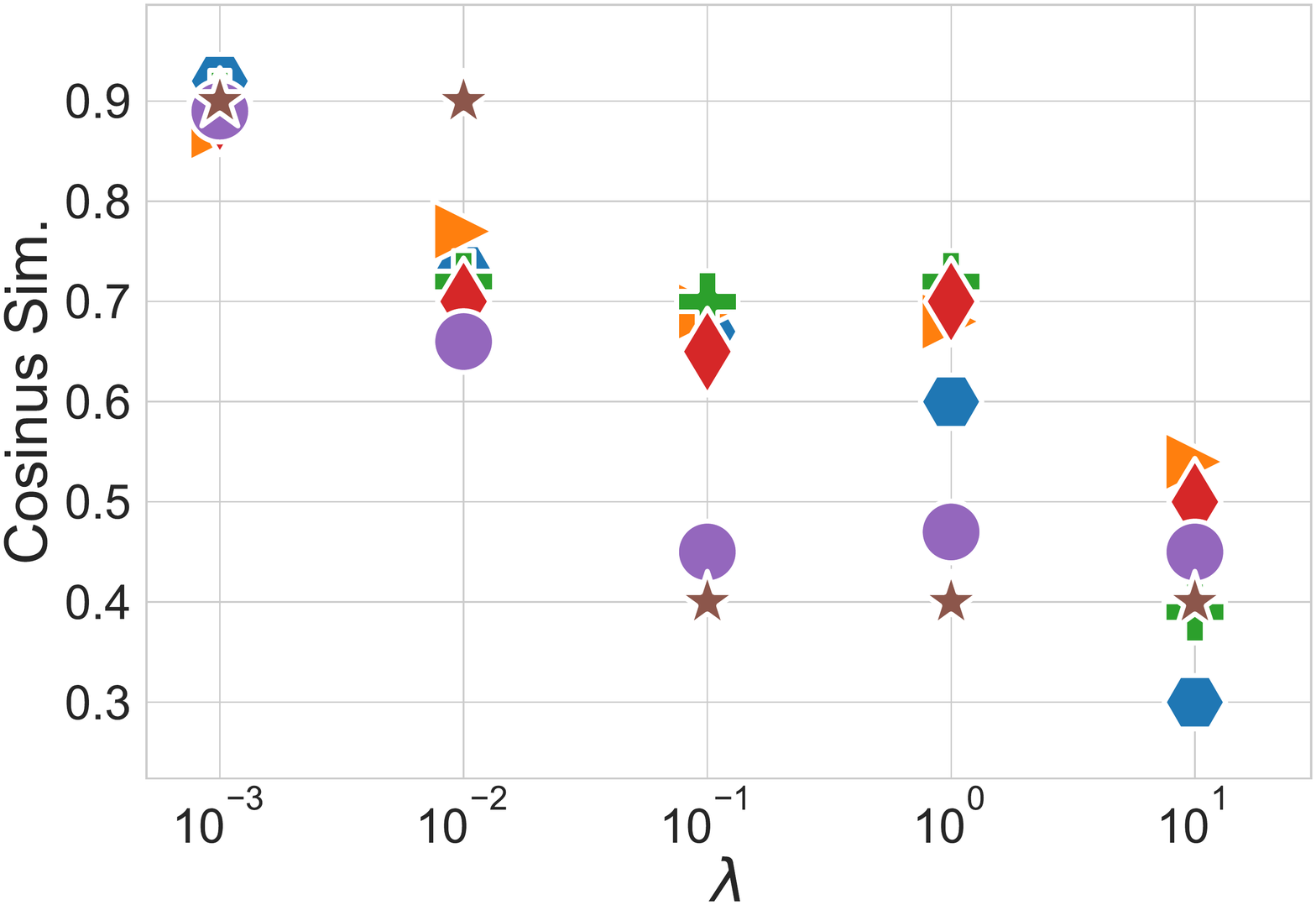}
    \end{subfigure}
        \caption{Results of cosine similarity on multiclass style transfer using categorical labels}\label{fig:cosin_category_st}
\end{figure}

\section{Additional Details on the Model}\label{sec:additionnal_styleEmb}
\subsection{Baseline Schemas}
We report in \autoref{fig:baseline_system} the schema of the proposed approach as well as the baselines.
\begin{figure*}
\centering     \begin{subfigure}[t]{.8\textwidth}
        \includegraphics[width=\textwidth]{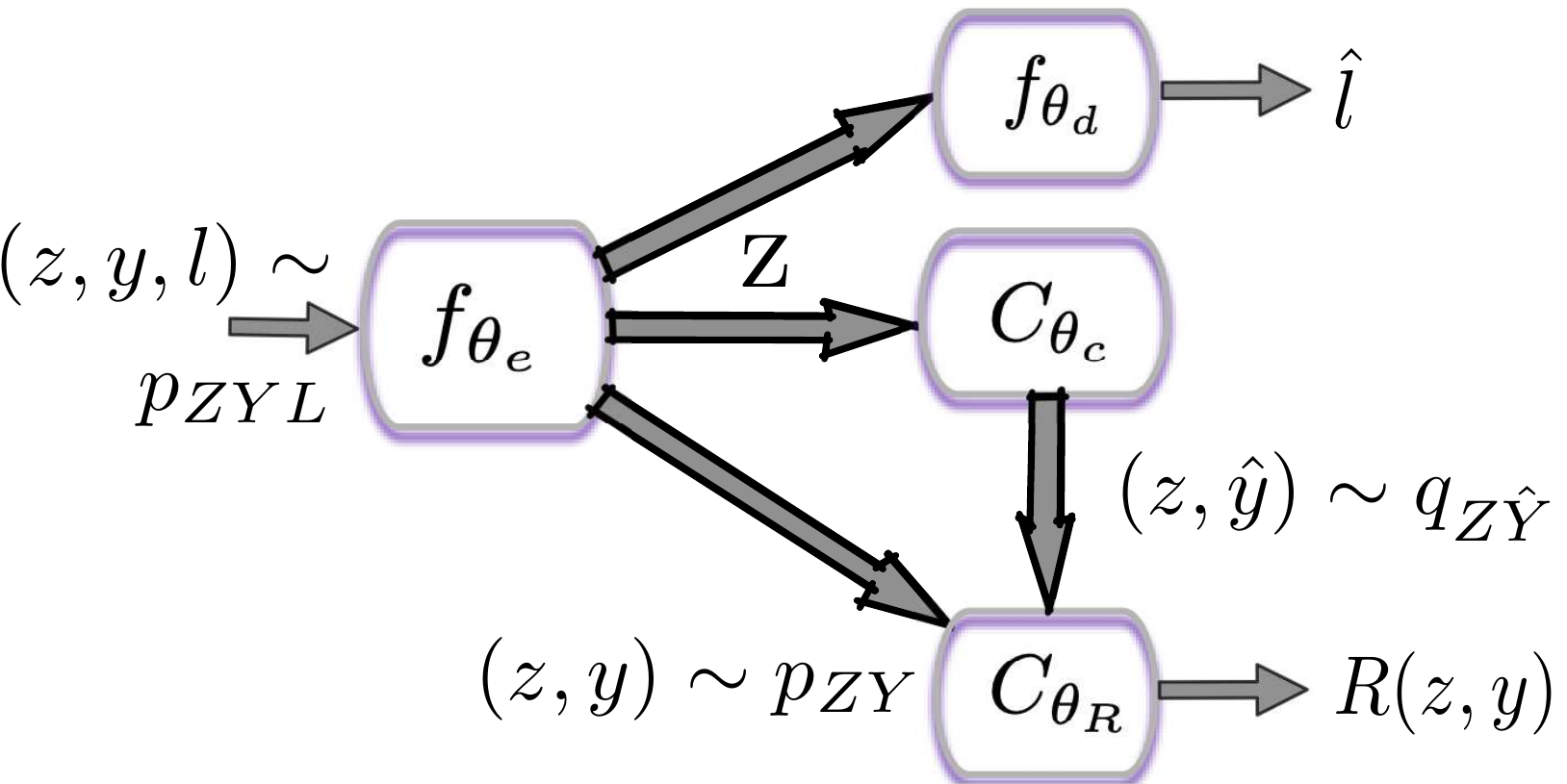}
        \caption{Classifier with our MI surrogate}  \vspace{1cm}
    \end{subfigure} 
 
\begin{subfigure}[t]{.8\textwidth}
        \includegraphics[width=\textwidth]{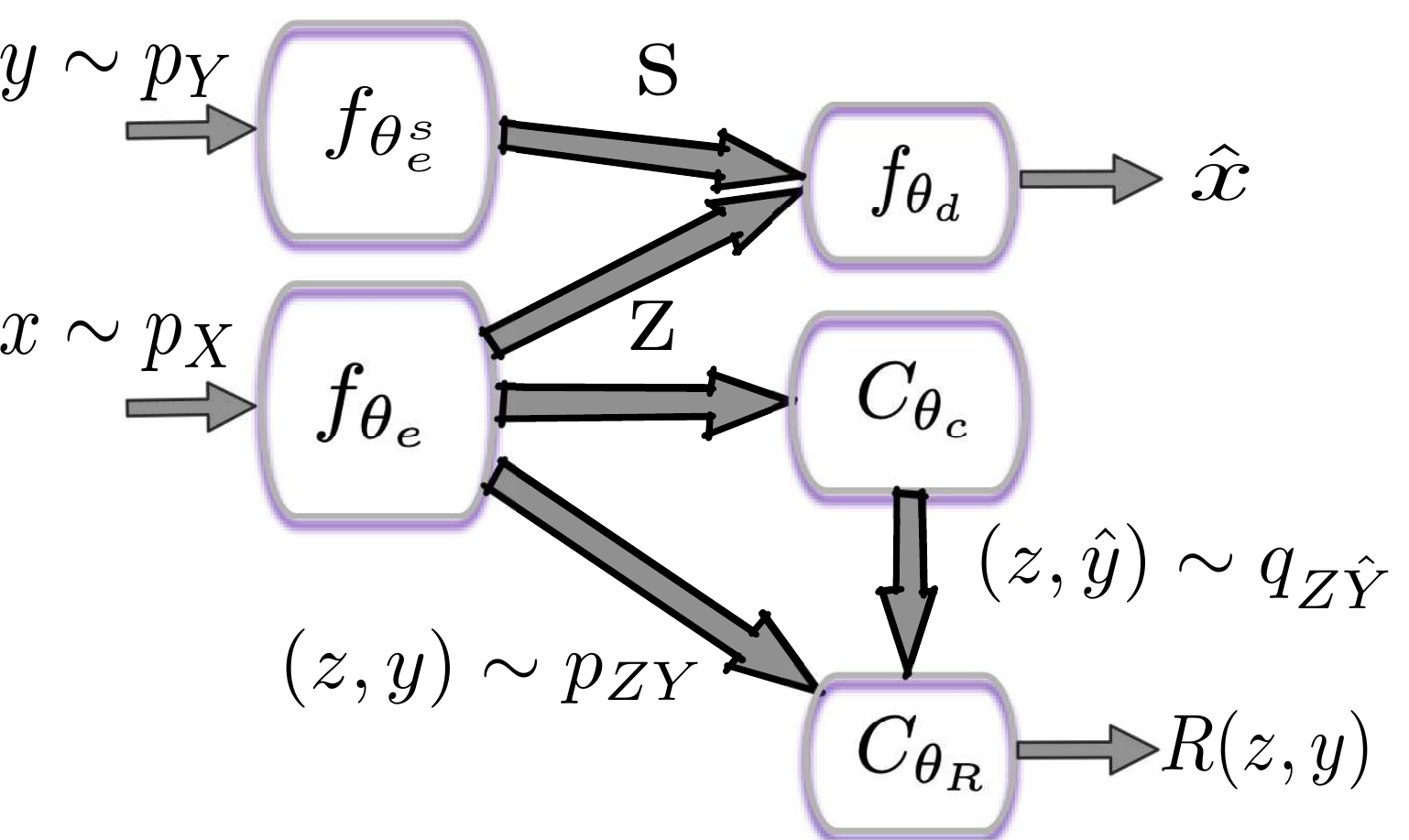}
        \caption{StyleEmb model from \cite{text} with our MI surrogate}\label{fig:baseline_emb} 
    \end{subfigure} \\
        \caption{Proposed methods. As described in \autoref{th:one}.}\label{fig:baseline_system}
\end{figure*}
\begin{figure*}
\centering     \begin{subfigure}[t]{.8\textwidth}
        \includegraphics[width=\textwidth]{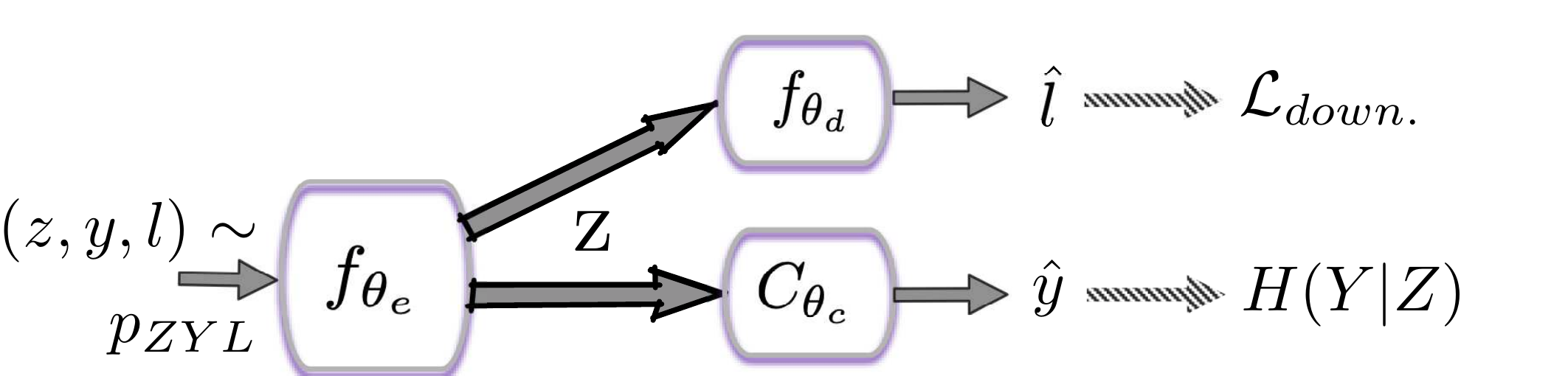}
        \caption{Classifier with adversarial loss from \cite{adversarial_removal}}\label{fig:fair_classification_baseline}     \vspace{1cm}
    \end{subfigure} 
 
\begin{subfigure}[t]{.8\textwidth}
        \includegraphics[width=\textwidth]{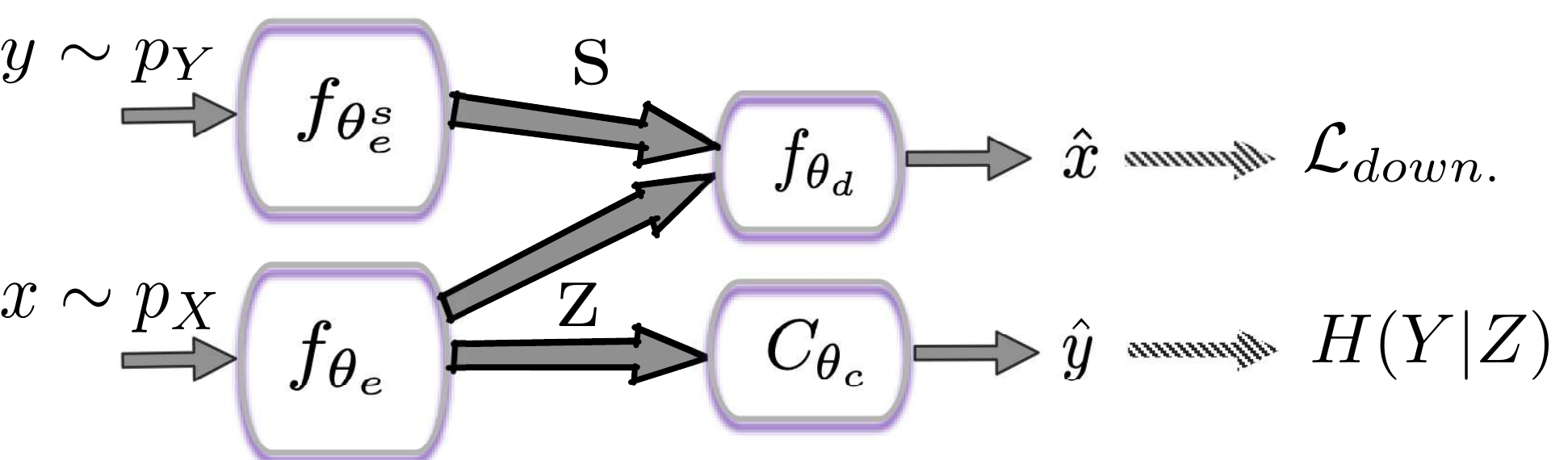}
        \caption{StyleEmb model from \cite{text}}\label{fig:baseline_emb} 
    \end{subfigure} \\
        \caption{Baselines methods, theses models use an adversarial loss for disentanglement. $f_{\theta_e}$ represents the input sentence encoder; $f_{\theta_e}^s$ denotes the style encoder (only used for sentence generation tasks); $C_{\theta_c}$ represents the adversarial classifier; $f_{\theta_d}$ represents the decoder that can be either a classifier (\autoref{fig:fair_classification_baseline} or a sequence decoder (\autoref{fig:baseline_emb}). Schemes of our proposed models are given in \autoref{fig:system} }\label{fig:baseline_system}
\end{figure*}

\subsection{Architecture Hyerparameters}
We use an encoder parameterized by a 2-layer bidirectional GRU \cite{gru} and a 2-layer decoder GRU. Both GRU and our word embedding lookup tables, trained from scratch, and have a dimension of 128 (as already reported by \cite{garcia2019token}, building experiments on  higher dimensions produces marginal improvement). The style embedding is set to a dimension of 8. The attribute classifier are MLP and are composed of 3 layer MLP with 128 hidden units and LeakyReLU \cite{leaky} activations, the dropout \cite{dropout} rate is set to 0.1. All models are optimised with AdamW \cite{adam,adamw} with a learning rate of $10^{-3}$ and the norm is clipped to $1.0$. Our model's hyperparameters have been set by a preliminary training on each downstream task: a simple classifier for the fair classification and a vanilla seq2seq \cite{seq2seq,colombo2020guiding} for the conditional generation task. The models requested for the classification task are trained during $100k$ steps while 300k steps are used for the generation task.

\section{Additional Details on the experimental Setup}\label{sec:addition_metric}
In this section, we provide additional details on the metric used for evaluating the different models.
\subsection{Content Preservation: BLEU \& Cosine Similarity}
Content preservation is an important aspect of both conditional sentence generation and style transfer. We provide here the implementation details regarding the implemented metrics.

\textbf{BLEU}.  For computing the BLEU score we choose to use the corpus level method provided in python sacrebleu \cite{sacrebleu} library \url{https://github.com/mjpost/sacrebleu.git}. It produces the official WMT scores while working with plain text.

\textbf{Cosine Similarity}. For the cosine similarity, we follow the definition of \citet{text} by taking the cosinus between source and generated sentence embedding. For computing the embedding we rely on the bag of word model and take the mean pooling of word embedding. We choose to use the pre-trained word vectors provided in \url{https://fasttext.cc/docs/en/pretrained-vectors.html}. They are trained on Wikipedia using fastText. These vectors in dimension 300 were obtained using the skip-gram model described in \citet{fasttext,fastext_1} with default parameters.

\subsection{Fluency: Perplexity}
To evaluate fluency we rely on the perplexity \cite{jalalzai2020heavy}, we use GPT-2 \cite{gpt} fine-tuned on the training corpus. GPT-2 is pre-trained on the BookCorpus dataset \cite{bookcorpus} (around 800M words).  The model has been taken from the HuggingFace Library \cite{hf}. Default hyperparameters have been used for the finetuning.

\subsection{Style Conservation/Transfer}
For style conservation \cite{colombo2019affect} (\textit{e.g.},  polarity, gender or category) we train a fasttext \cite{fasttext,fastext_2,fastext_1} classifier \url{https://fasttext.cc/docs/en/supervised-tutorial.html}. We use the validation corpus to select the best model. Preliminary comparisons with deep classifiers (based on either convolutionnal layers or recurrent layers) show that fasttext obtains similar result while being litter and faster.

\subsection{Disentanglement}
For disentanglement, we follow common practice \cite{multiple} and implement a two layers perceptron \cite{perceptron}. We use LeakyRelu \cite{leaky} as activation functions and set the dropout \cite{dropout} rate to $0.1$.

\section{Additional Results on Sentiment}\label{sec:addition_sentiment}
\subsection{Binary Sentence Generation}


\subsubsection{Human Evaluation} 
In \autoref{tab:human_annotation}, we report the performances of systems when evaluated by humans on the polarity transfer task. 100 sentences are generated by each system and 3 english native speakers are asked to annotate each sentence along 3 dimensions (\textit{i.e} fluency, sentiment and content preservation). Turkers assign binary labels to fluency and sentiment (following the protocol introduced in \citet{jalalzai2020heavy}) while content is evaluated on a likert scale from 1-5. For content preservation, both the input sentence and the generated sentence are provided to the turker. The annotator agreement is measure by the Krippendorff Alpha\footnote{Krippendorff Alpha measures of inter-rater reliability in $[0,1]$: $0$ is perfect disagreement and $1$ is perfect agreement.} \cite{krippendorff2018content}. The Krippendorff Alpha is: $\alpha=0.54$ on the sentiment classification, $\alpha=0.20$ for fluency and $\alpha=0.18$ for content preservation.

\begin{table}[t]
\centering
   \begin{tabular}[h]{c|ccc}
   \hline 
   Model &  Fluency & Content & Sentiment\\\hline
Human& 0.80 & 3.4 & 0.78 \\\hline
$Adv$ & 0.60& 2.4 & 0.63 \\
$vCLUB-S$ & 0.62& 2.6 & 0.65 \\
$KL$  & 0.68 & 2.6 &  0.63 \\
$D_{\alpha = 1.3}$ & 0.70 & 2.4 &  0.65  \\
$D_{\alpha = 1.5}$ &0.68 & 2.9 &  0.70  \\
$D_{\alpha = 1.8}$ &0.76& 3.0 &  0.58 \\\hline
\end{tabular}
\caption{Human annotation of generated samples. For this comparison we rely on the sentences provided in \url{https://github.com/rpryzant/delete_retrieve_generate}. Human annotations are also provided by \citet{li2018delete}. We have reprocessed the provided sentence using a tokenizer based on SentencePiece \cite{tok_0,tok_1}. Since there is a trade-off between automatic evaluation metrics (\textit{i.e} BLEU, Perplexity and Accuracy of Style Transfer), we set
minimum thresholds on BLEU and on style transfert accuracy. The best model that met the threshold on validation is selected. We will release--along with our code--new generated sentences for comparison.}
\label{tab:human_annotation}
\end{table}
\subsection{Content preservation using Cosine Similarity}
\begin{figure} 

\centering     
        \includegraphics[width=0.45\textwidth]{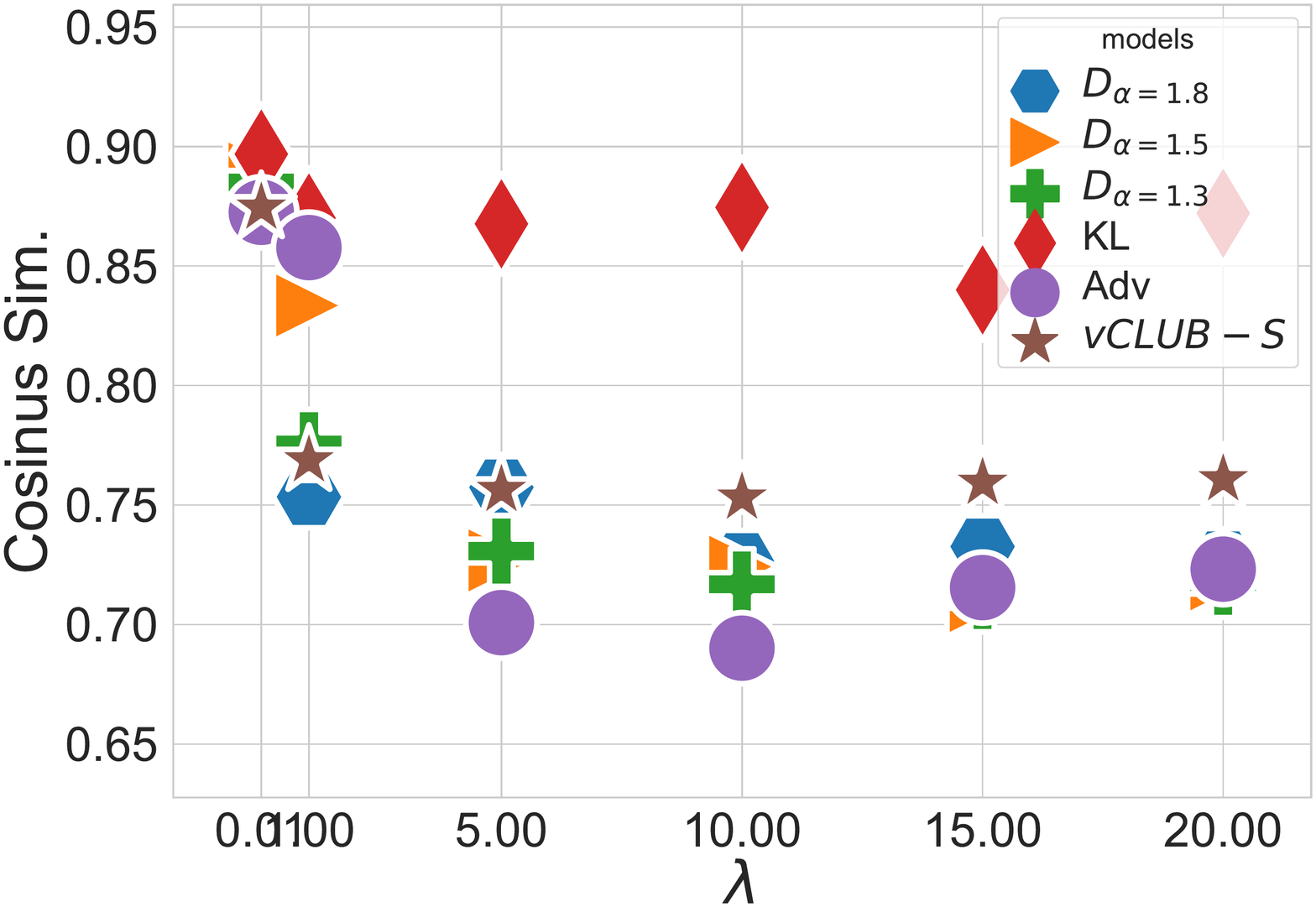}
        \caption{Content preservation measured by the cosine similarity. 
        }\label{fig:add_style_transfert_sentiment}
\end{figure} 

\autoref{fig:add_style_transfert_sentiment} measures the content preservation measured using cosine similarity for the sentence generation task using sentiment labels. As with the BLEU score, we observe that as the learnt representation becomes more entangled ($\lambda$ increases) less content is preserved. Similarly to BLEU the model using the KL bound conserves outperforms other models in terms of content preservation for $\lambda > 5$.

\subsection{Example of generated sentences}
\autoref{tab:example_sentences_st} 
gathers some sentences generated by the different sentences for different values of $\lambda$. 

\textbf{Style transfert.} From \autoref{tab:example_sentences_st}, we can observe that the impact of disentanglement on a qualitative point of view. For small values of $\lambda$ the models struggle to do the style transfer (see example 2 for instance). As $\lambda$ increases disentanglement becomes easier, however, the content becomes more generic which is a known problem (see \cite{mmi} for instance). 


\textbf{Example of ``degeneracy" for large values of $\lambda$.} For sentences generated with the baseline model a repetition phenomenon appears for greater values of $\lambda$. For certain sentences, models ignore the style token (\textit{i.e.}, the sentence generated with a positive sentiment is the same as the one generated with the negative sentiment). We attribute this degeneracy to the fact that the model is only trained with $(x_i,y_i)$ sharing the same sentiment which appears to be an intrinsic limitation of the model introduced  by \cite{text}.

\textbf{Analysis of performances of vCLUB-S} Similarly to what can be observed with automatic evaluation 
\autoref{tab:example_sentences_st} shows that the system based on vCLUB-S has only two regimes: “light” disentanglement and strong disentanglement. With light disentanglement the decoder fail at transferring the polarity and for strong disentanglement few content features remain and the system tends to output generic sentences.

\section{Additional Results on Multi class Sentence Generation}\label{sec:addition_multu}
Results on the multi-class style transfer and on 
are reported in \autoref{fig:accuracy_category_st} 
Similarly than in the binary case there exists a trade-off between content preservation and style transfer accuracy. We observe that the BLEU score in this task is in a similar range than the one in the gender task, which is expected because data come from the same dataset where only the labels changed.

\begin{table}
\resizebox{0.5\textwidth}{!}{\begin{tabular}{ll|l} 
\hline
$\lambda$ & Model	& Sentence	\\\hline
\multirow{7}{*}{0.1} &\textbf{Input}  &  	\textbf{It's freshly made, very soft and flavorful.} \\\hline
&Adv  & it's crispy and too nice and very flavor.  \\
&vCLUB-S  & It's freshly made, and great. \\
&KL  & it's a huge, crispy and flavorful.  \\
&$D_{\alpha=1.3}$  &  it's hard, and the flavor was flavorless. \\
&$D_{\alpha=1.5}$  & it's very dry and not very flavorful either.  \\
&$D_{\alpha=1.8}$  & it's a good place for lunch or dinner.  \\ \hline
\multirow{7}{*}{1} &Input  & 	it's freshly made, very soft and flavorful.  \\\hline
&Adv  &  it's not crispy and not very flavorful flavor. \\
&vCLUB-S  & It's bad. \\
&KL  & 	it's very fresh, and very flavorful and flavor.  \\
&$D_{\alpha=1.3}$  & it's not good, but the prices are good.  \\
&$D_{\alpha=1.5}$  &  it's not very good, and the service was terrible. \\
&$D_{\alpha=1.8}$ &	it was a very disappointing experience and the food was awful.   \\\hline
\multirow{7}{*}{5} &Input  & 	it's freshly made, very soft and flavorful.  \\\hline
&Adv  &  	i hate this place. \\
&vCLUB-S  & i hate it. \\
&KL  &  	it's very fresh, flavorful and flavorful. \\
&$D_{\alpha=1.3}$  &  	it's not worth the money, but it was wrong. \\
&$D_{\alpha=1.5}$  &  it's not worth the price, but not worth it. \\
&$D_{\alpha=1.8}$  & 	it's hard to find, and this place is horrible.  \\\hline
\multirow{7}{*}{10} &Input  & 	it's freshly made, very soft and flavorful.  \\\hline
&Adv  & i hate this place.  \\
&vCLUB-S  & i hate it. \\
&KL  & 	it's a little warm and very flavorful flavor.  \\
&$D_{\alpha=1.3}$  & it was a little overpriced and not very good.  \\
&$D_{\alpha=1.5}$  &  	it's a shame, and the service is horrible. \\
&$D_{\alpha=1.8}$  & 	it's not worth the \$ NUM.  \\\hline\hline
\end{tabular}}
\caption{
Sequences generated by the different models on the binary sentiment transfer task.}
\label{tab:example_sentences_st}
\end{table} 
\twocolumn
\section{Binary Sentence Generation: Application to Gender Data}\label{sec:addition_gender}
\subsection{Quality of the Disentanglement}

\begin{figure}
    \centering
    \includegraphics[width=0.5\textwidth]{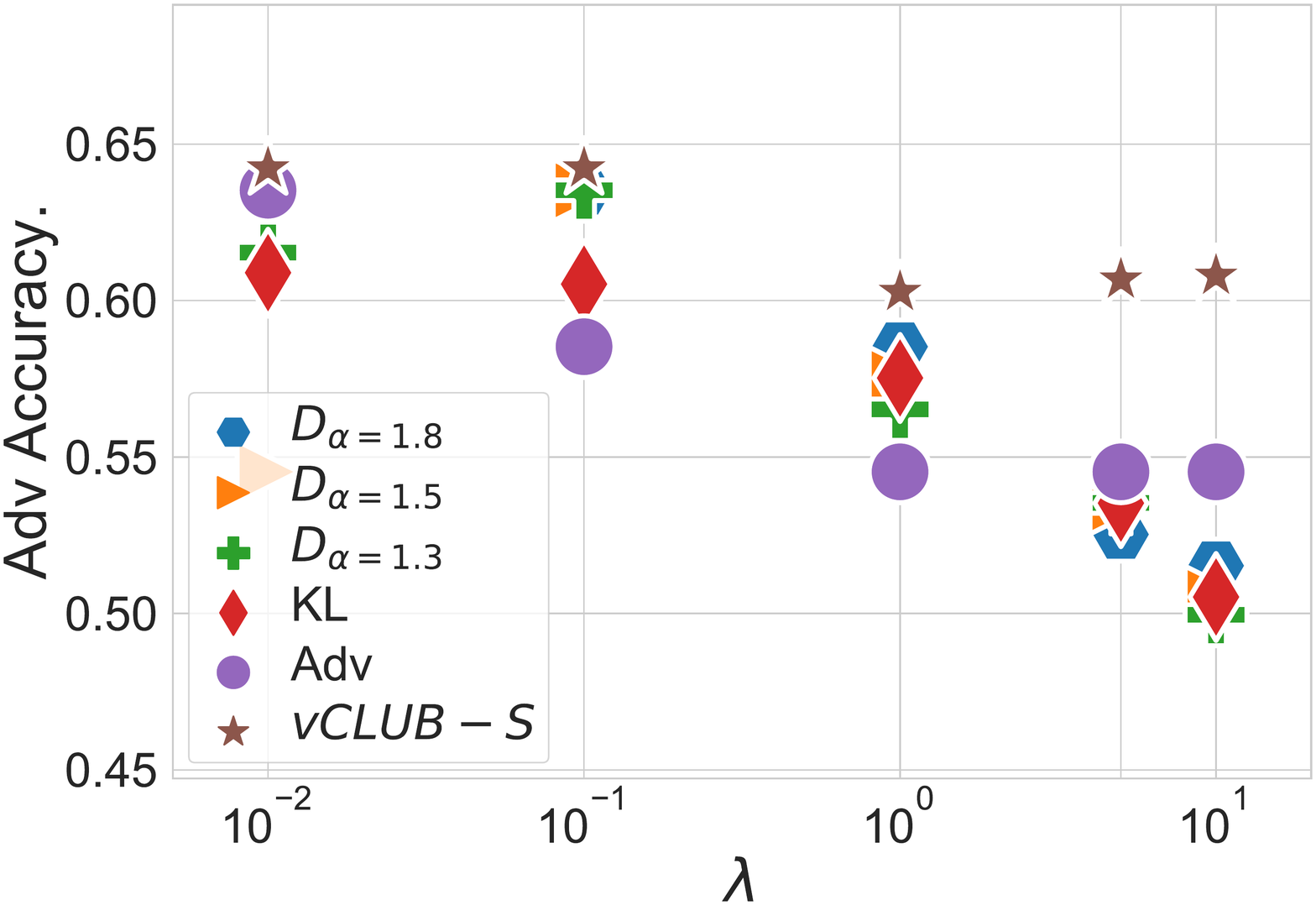}
    \caption{Disentanglement of the learnt embedding when training an off-line adversarial classifier for the sentence generation with gender data.}\label{fig:disent_gender_st}
\end{figure}

In \autoref{fig:disent_gender_st}, we report the adversary accuracy of the different methods for the values of $\lambda$. It is worth noting that gender labels are noisier than sentiment labels \cite{multiple}. We observe that the adversarial loss saturates at $55\%$ where a model trained on MI bounds can achieve a better disentanglement. Additionally, the models trained with MI bounds allow better control of the desired degree of disentanglement. 

\subsection{Quality of Generated Sentences}
\begin{figure*}
\centering     \begin{subfigure}[t]{0.45\textwidth}
        \includegraphics[width=\textwidth]{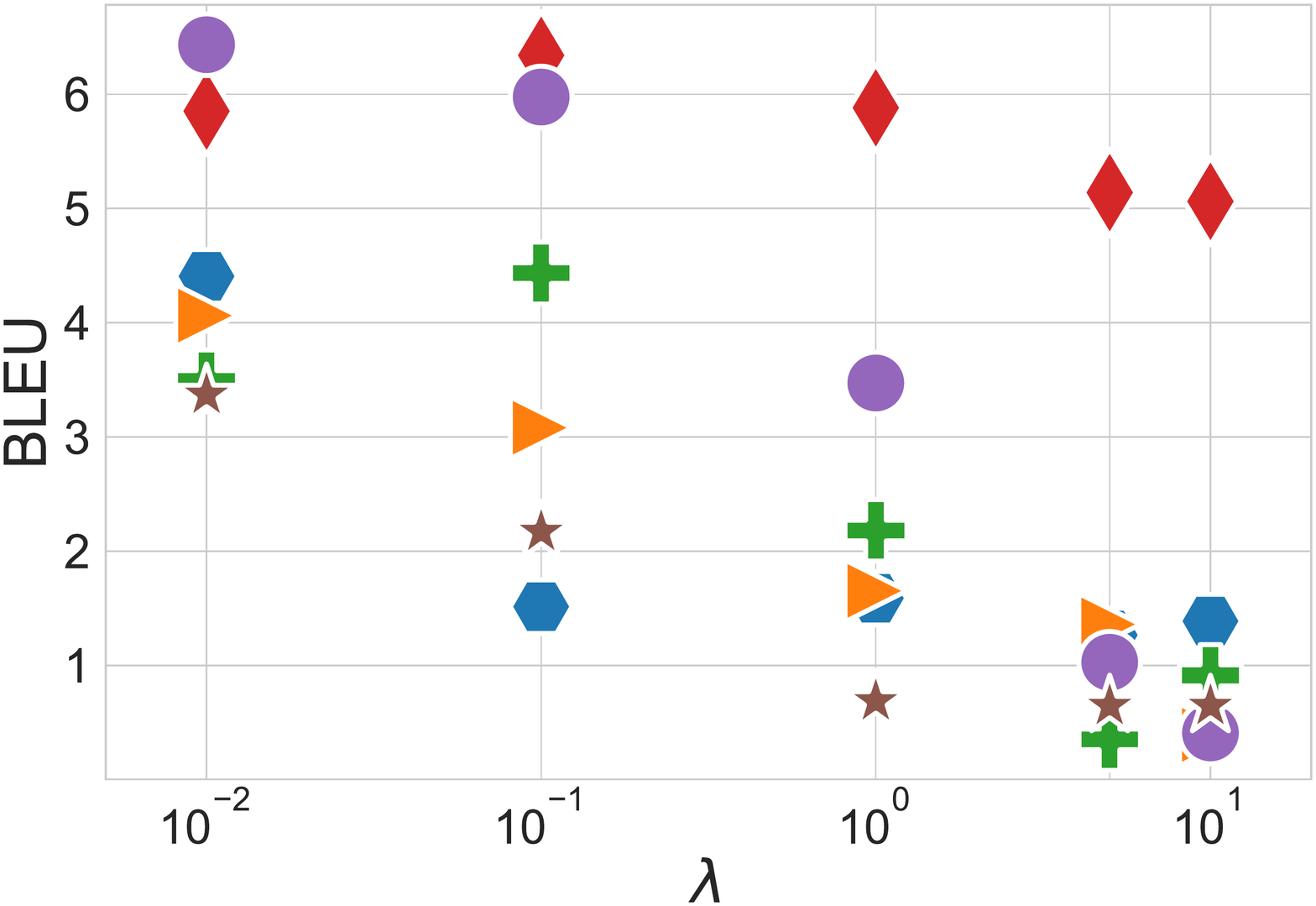}
        \caption{}\label{fig:bleu_gender_st}  
         
    \end{subfigure} 
\begin{subfigure}[t]{0.45\textwidth}
        \includegraphics[width=\textwidth]{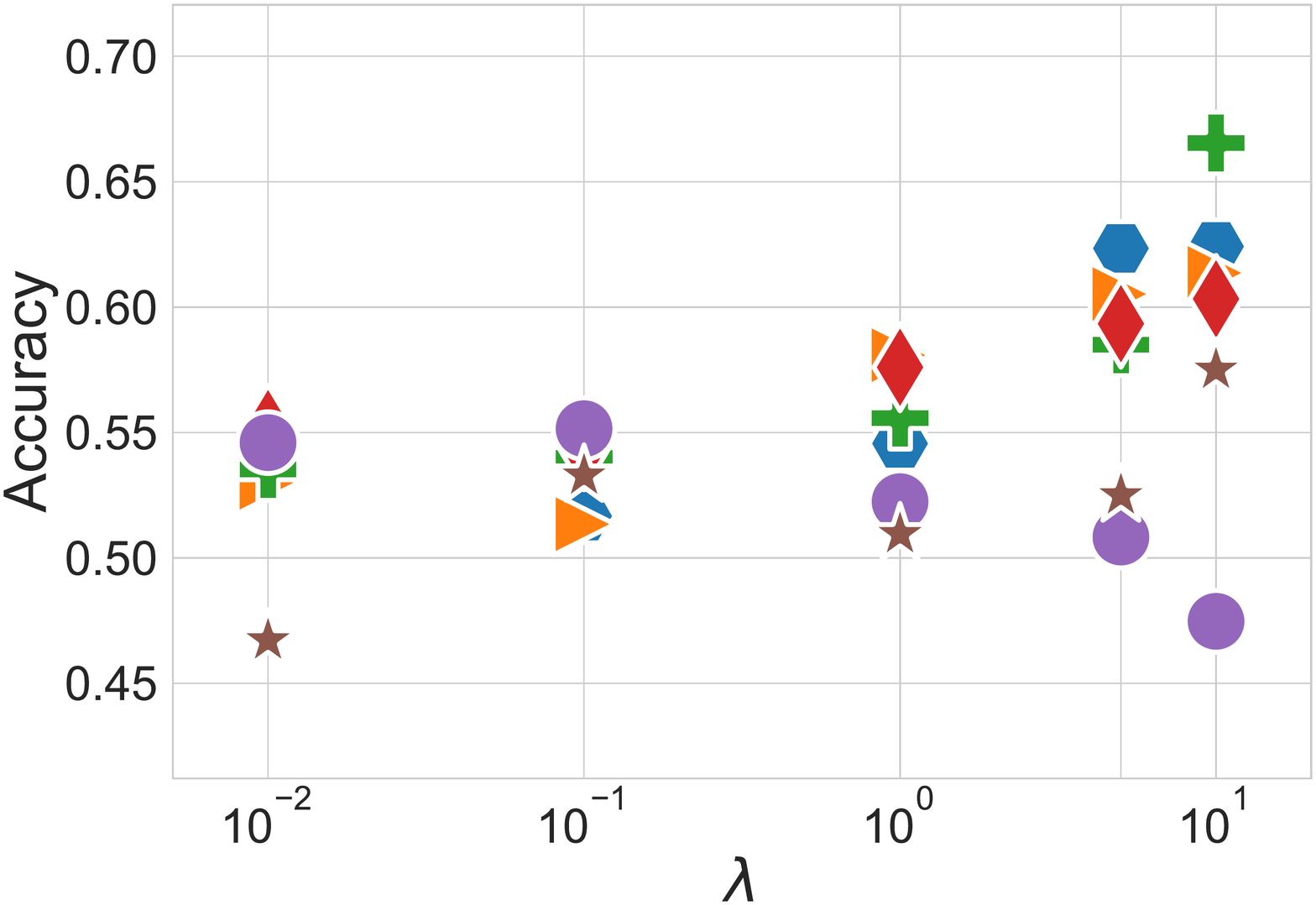}
        \caption{}\label{fig:accuracy_gender_st} 
    \end{subfigure} \\
    \begin{subfigure}[t]{0.45\textwidth}
        \includegraphics[width=\textwidth]{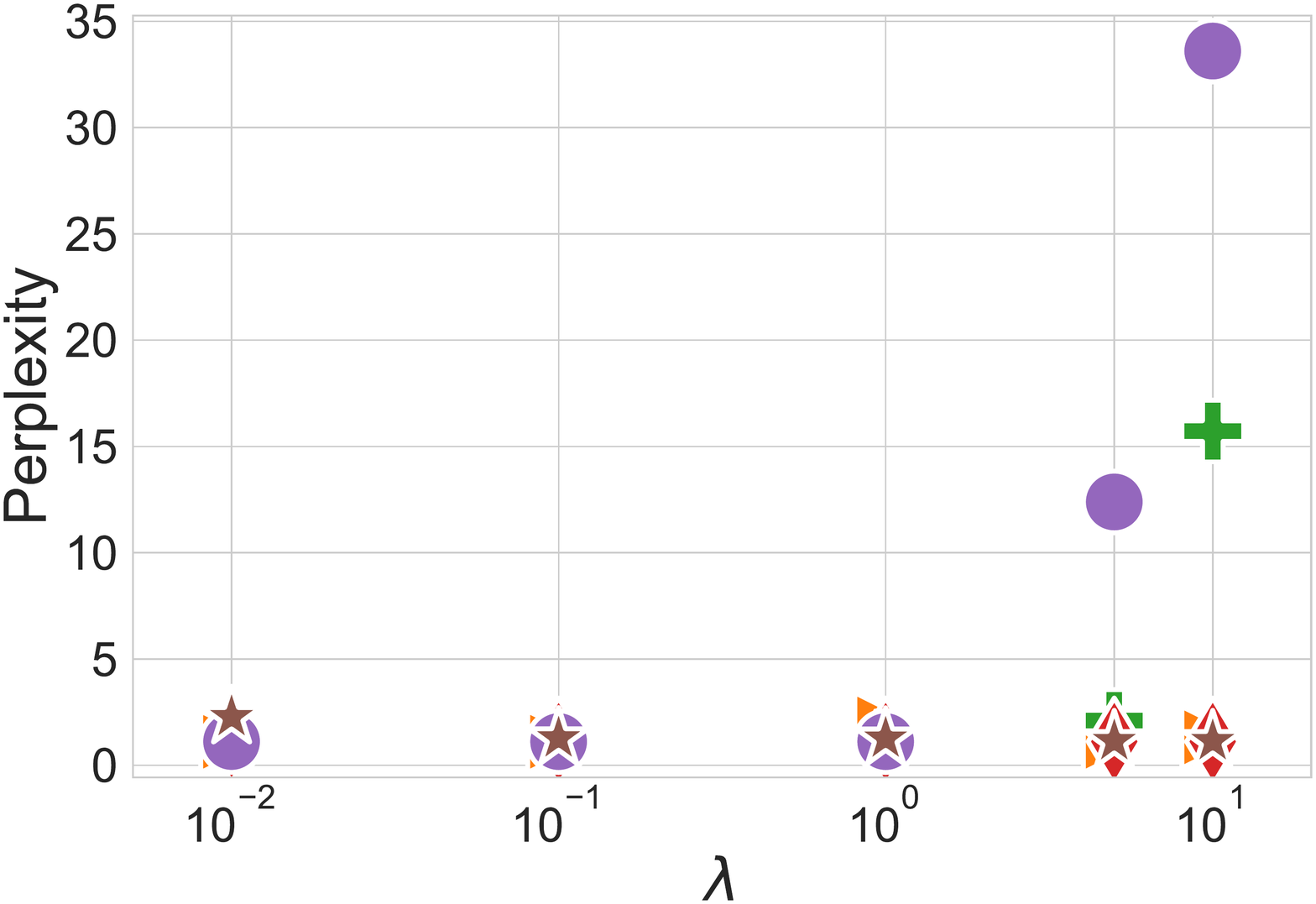}
         \caption{}\label{fig:ppl_gender_st}
    \end{subfigure}
    \begin{subfigure}[t]{0.45\textwidth}
        \includegraphics[width=\textwidth]{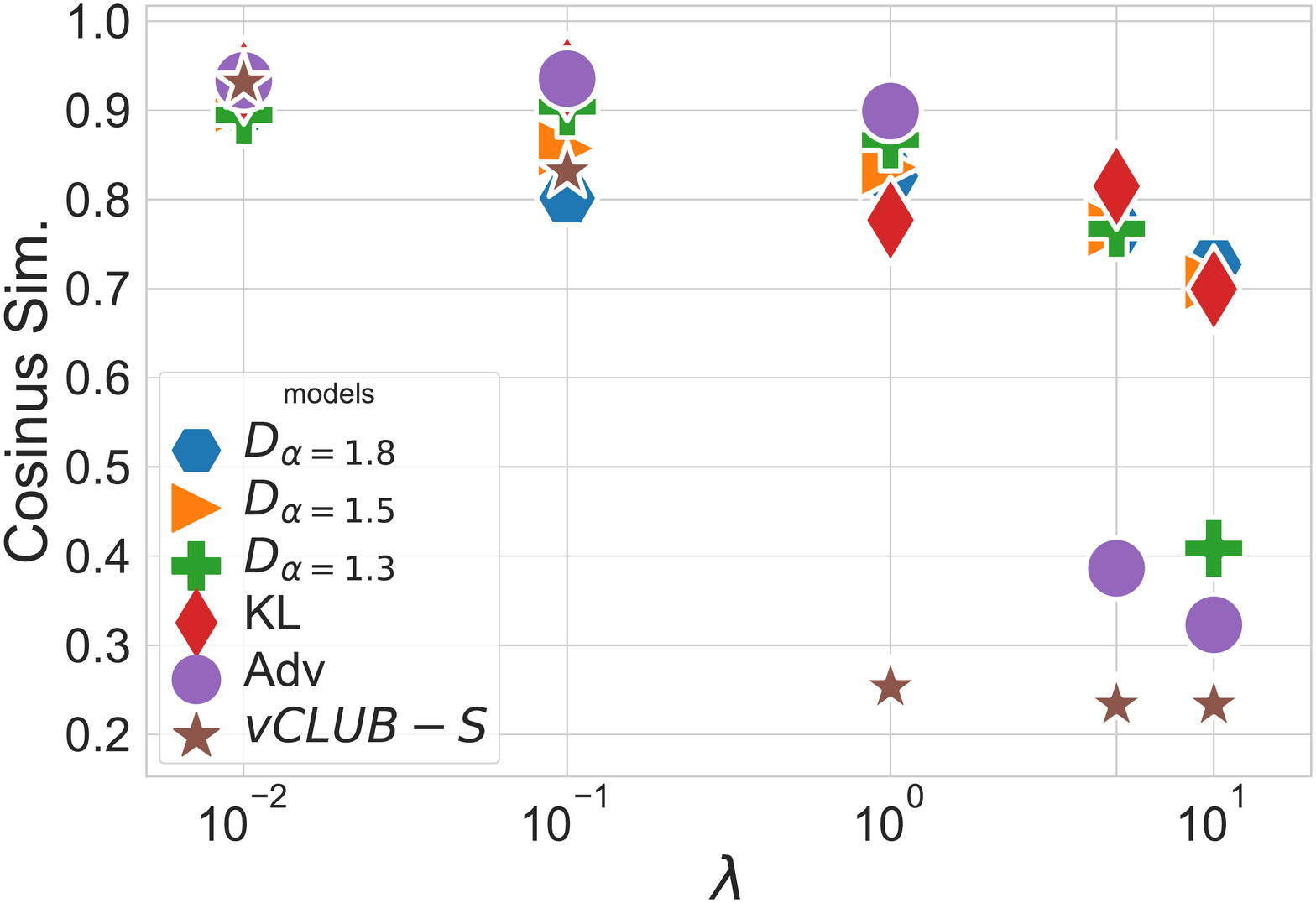}
         \caption{}\label{fig:cosim_gender_st}
    \end{subfigure}
        \caption{Numerical experiments on binary style transfer using gender labels. Results include: BLEU (\autoref{fig:bleu_gender_st}); cosine similarity (\autoref{fig:cosim_gender_st}); style transfer accuracy (\autoref{fig:accuracy_gender_st}); sentence fluency (\autoref{fig:ppl_gender_st}).}
        \label{fig:style_transfert_gender}
\end{figure*}

\begin{figure*}
\centering     \begin{subfigure}[t]{0.4\textwidth}
        \includegraphics[width=\textwidth]{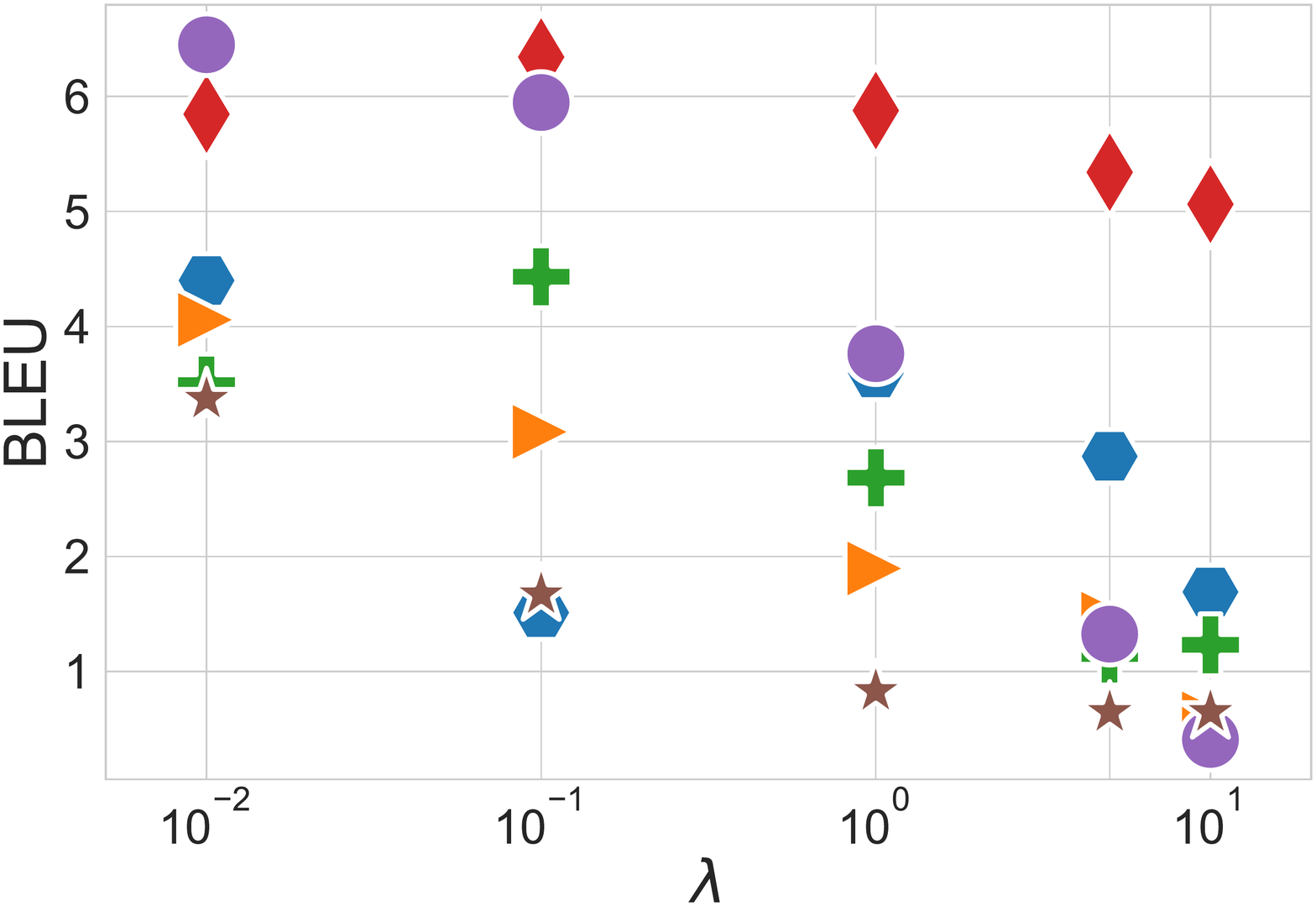}
        \caption{}\label{fig:bleu_gender_cg}  
    \end{subfigure} 
\begin{subfigure}[t]{0.4\textwidth}
        \includegraphics[width=\textwidth]{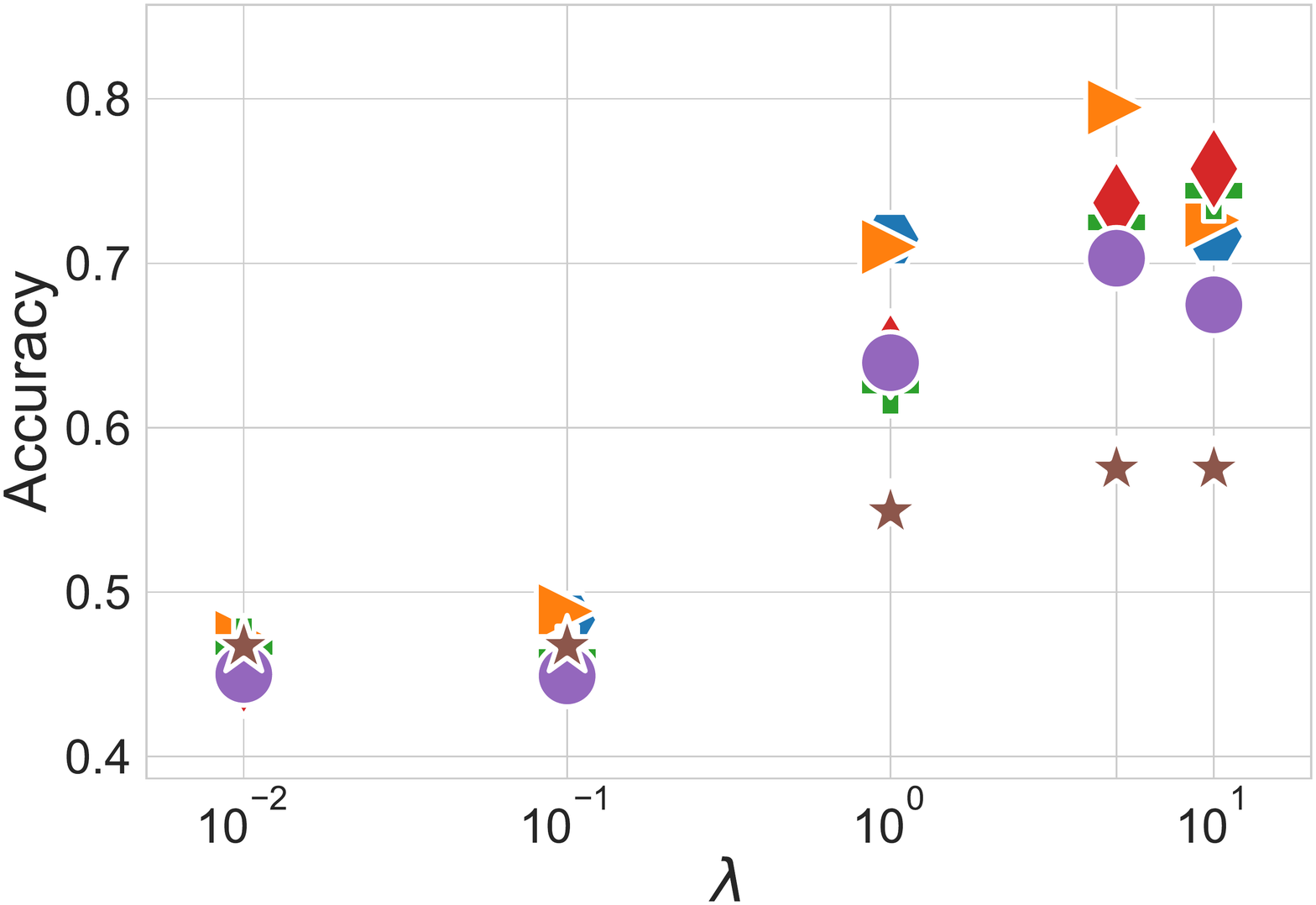}
        \caption{}\label{fig:accuracy_gender_cg} 
    \end{subfigure} 
    \begin{subfigure}[t]{0.4\textwidth}
        \includegraphics[width=\textwidth]{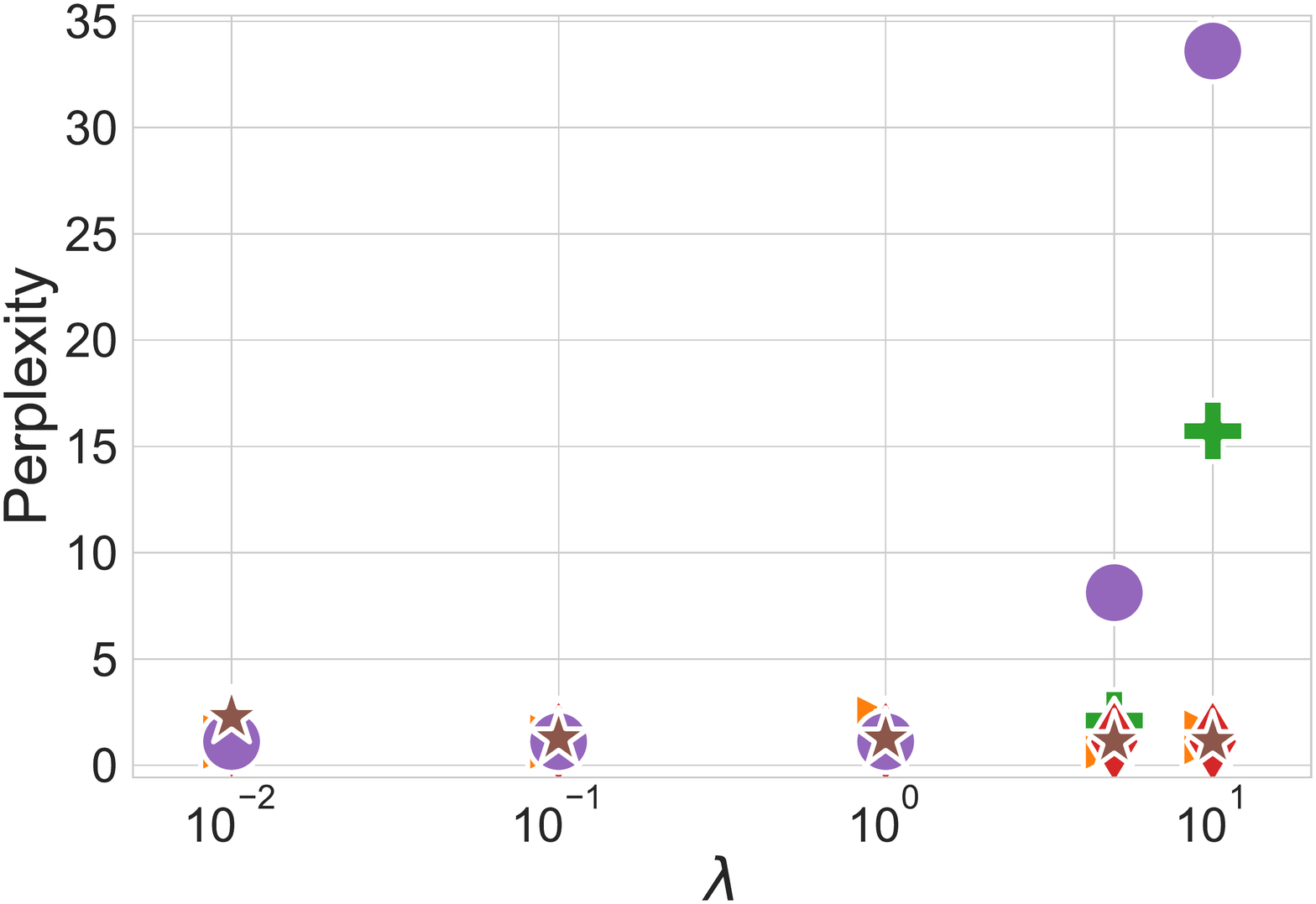}
         \caption{}\label{fig:ppl_gender_cg}
    \end{subfigure}
        \begin{subfigure}[t]{0.4\textwidth}
        \includegraphics[width=\textwidth]{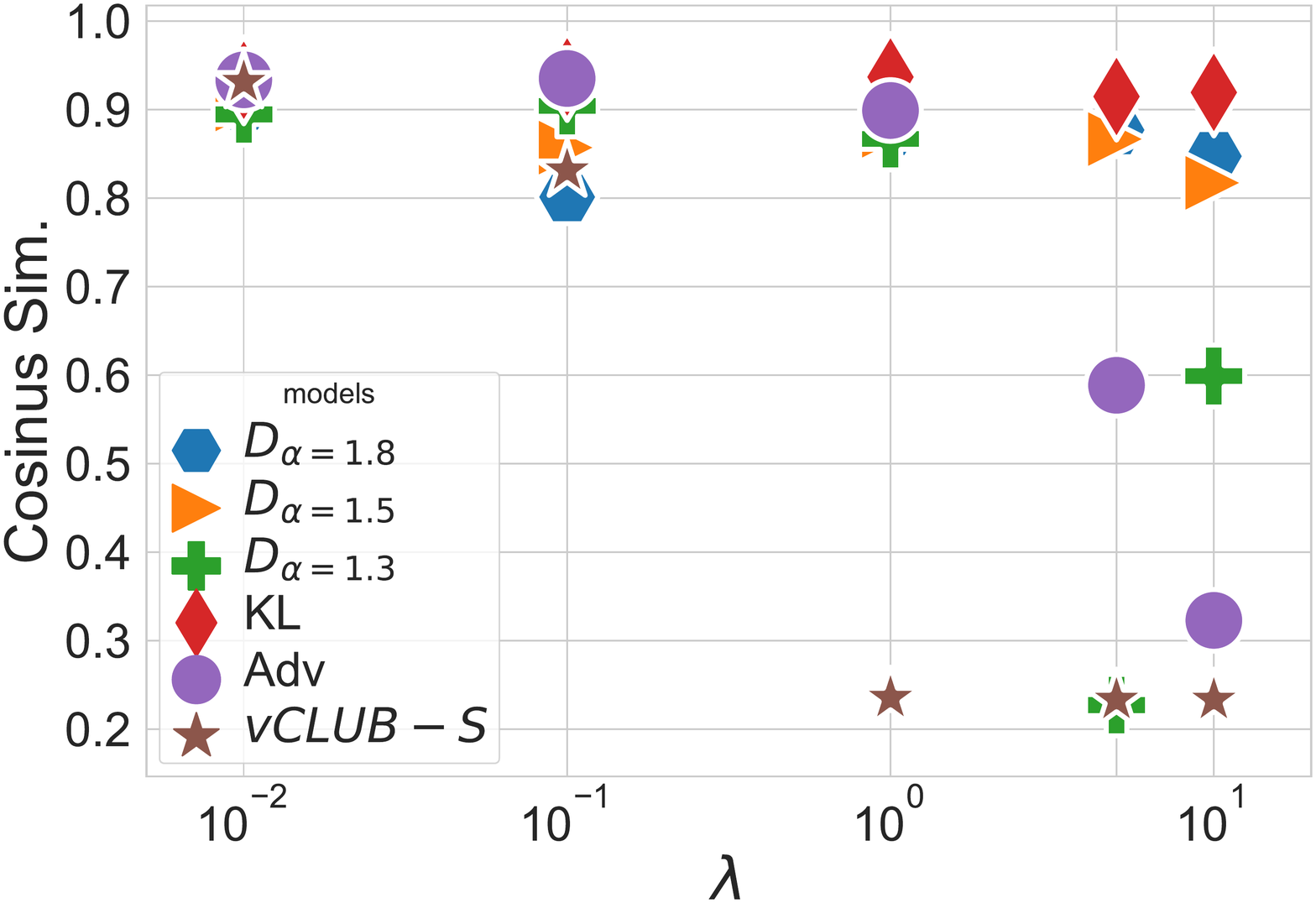}
         \caption{}\label{fig:cosim_gender_cg}
    \end{subfigure}
        \caption{Numerical experiments on conditional sentence generation using gender labels. Results includes: BLEU (\autoref{fig:bleu_gender_cg}); cosine similarity (\autoref{fig:cosim_gender_cg}); style transfer accuracy (\autoref{fig:accuracy_gender_cg}); sentence fluency (\autoref{fig:ppl_gender_cg}).}\label{fig:cg_gender}
\end{figure*}

Results on the sentence generation tasks are reported in \autoref{fig:style_transfert_gender} and in \autoref{fig:cg_gender}.
We observe that for $\lambda > 1$ the adversarial loss degenerates as observe in the sentiment experiments.Compared to sentiment score we observe a lower score of BLEU which can be explained by the length of the review in the FYelp dataset. On the other hand, we observe a similar trade-off between style transfer accuracy and content preservation in the non degenerated case: as style transfer accuracy increases, content preservation decreases. Overall, we remark a behaviour similar to the one we observe in sentiment experiments.

\section{Additional Results on Multi class Sentence Generation}\label{sec:addition_multu}

Results on the multi-class style transfer and on conditional sentence generation 
are reported in \autoref{fig:accuracy_category_st} and \autoref{fig:accuracy_sentiment_cg}. 
Similarly than in the binary case there exists a trade-off between content preservation and style transfer accuracy. We observe that the BLEU score in this task is in a similar range than the one in the gender task, which is expected because data come from the same dataset where only the labels changed.

\begin{figure*}
\centering     \begin{subfigure}[t]{0.4\textwidth}
        \includegraphics[width=\textwidth]{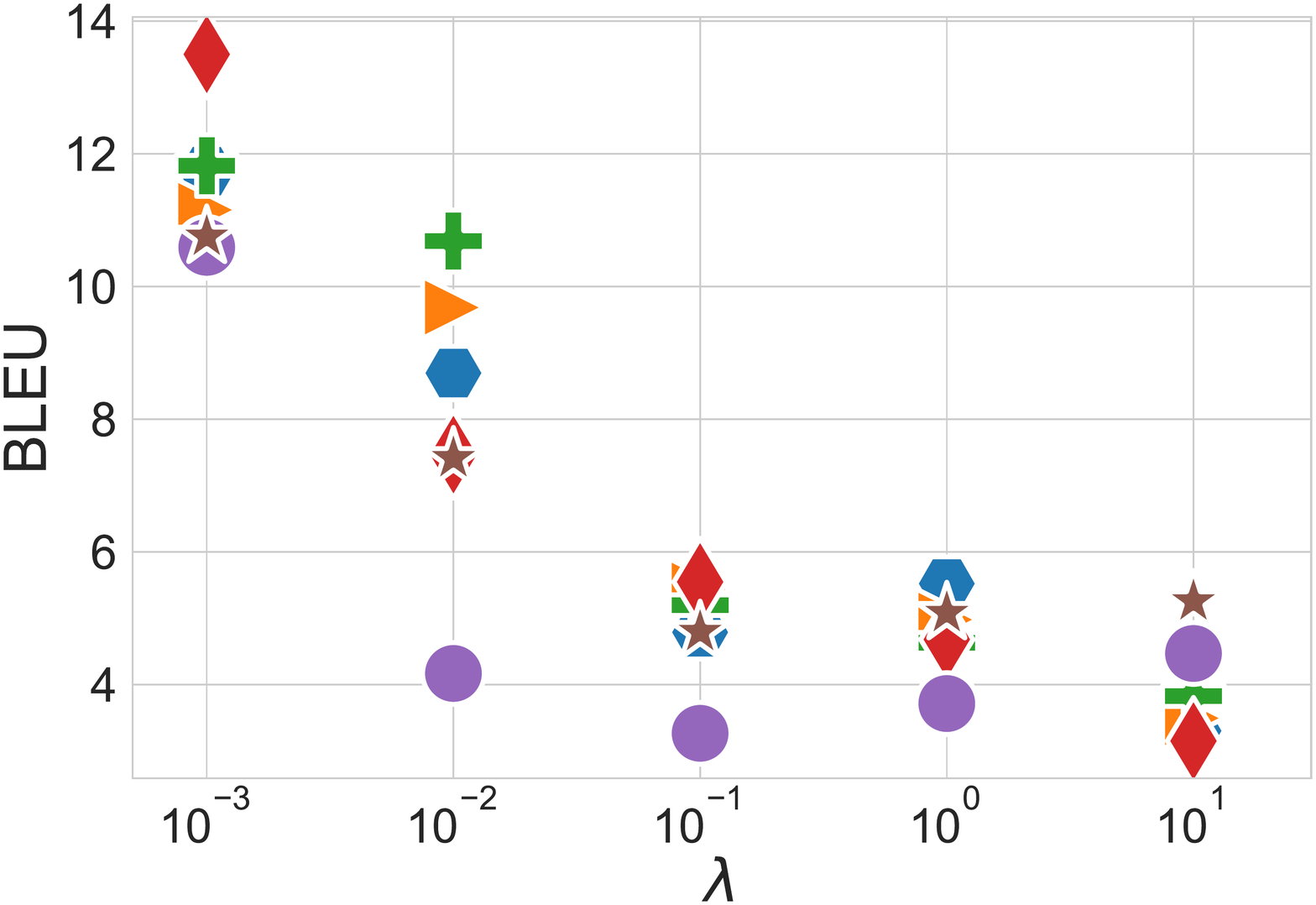}
        \caption{}\label{fig:bleu_category_cg}  
    \end{subfigure} 
\begin{subfigure}[t]{0.4\textwidth}
        \includegraphics[width=\textwidth]{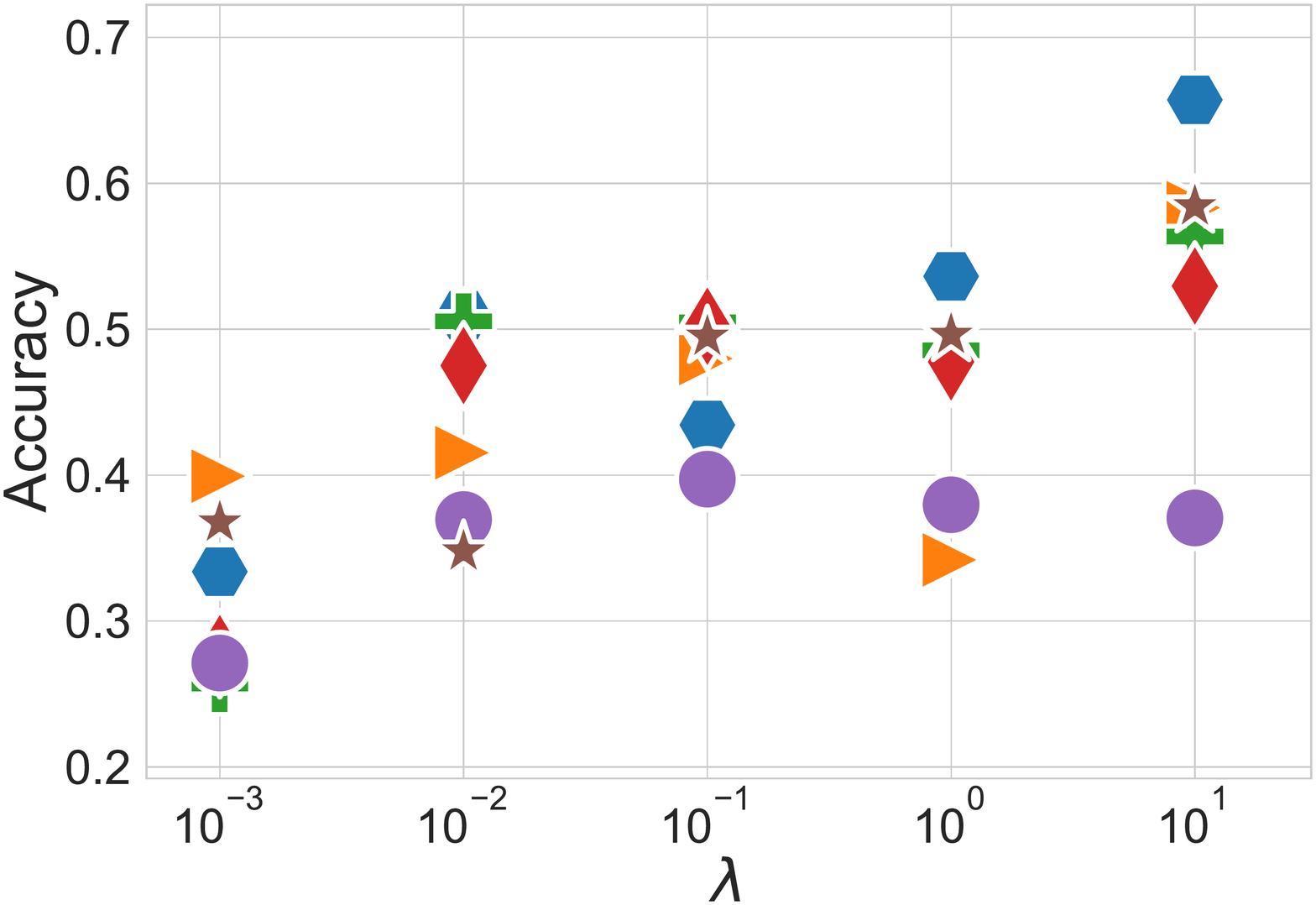}
        \caption{}\label{fig:accuracy_category_cg} 
    \end{subfigure} \begin{subfigure}[t]{0.4\textwidth}
        \includegraphics[width=\textwidth]{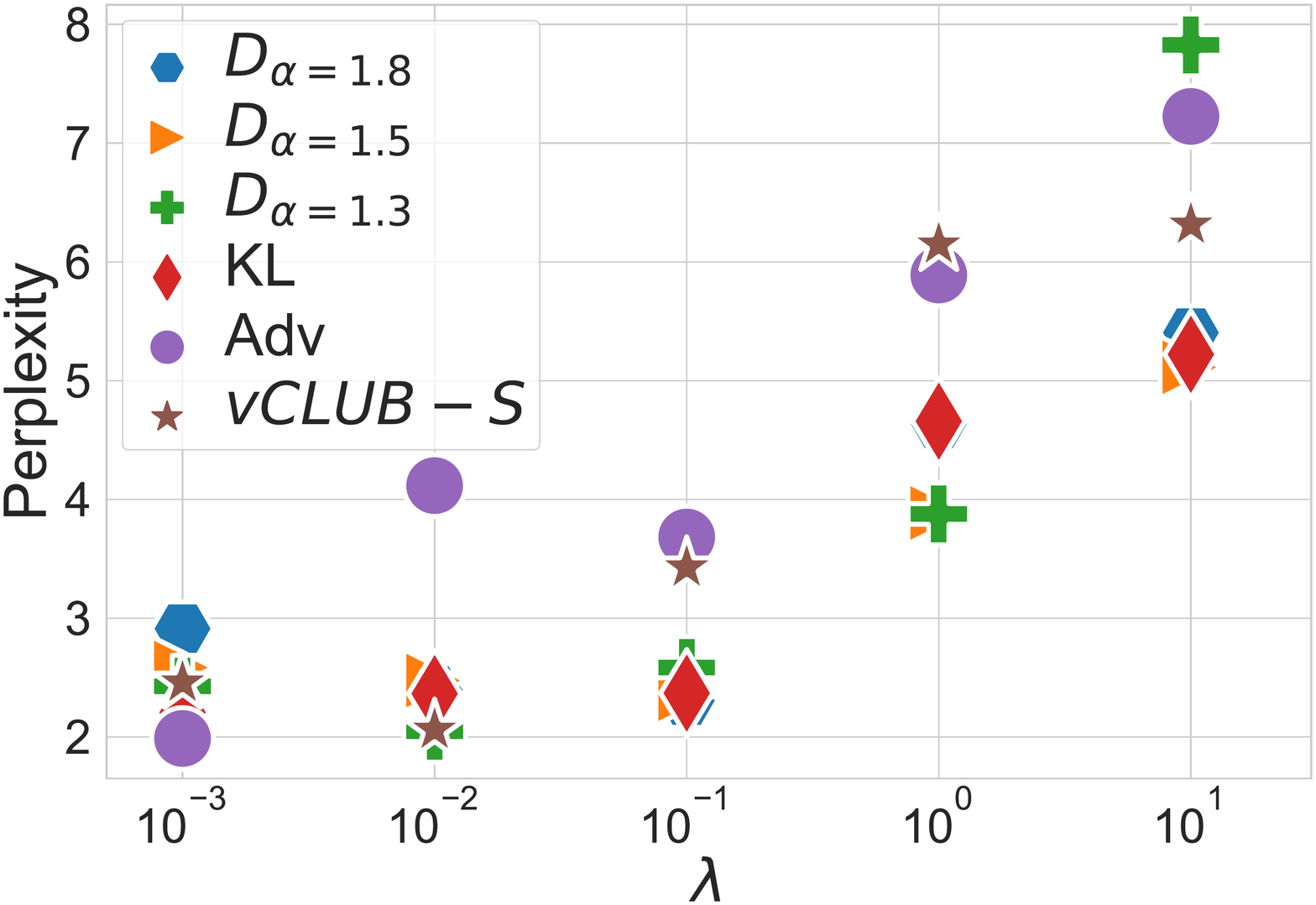}
         \caption{}\label{fig:ppl_category_cg}
    \end{subfigure}
        \begin{subfigure}[t]{0.4\textwidth}
        \includegraphics[width=\textwidth]{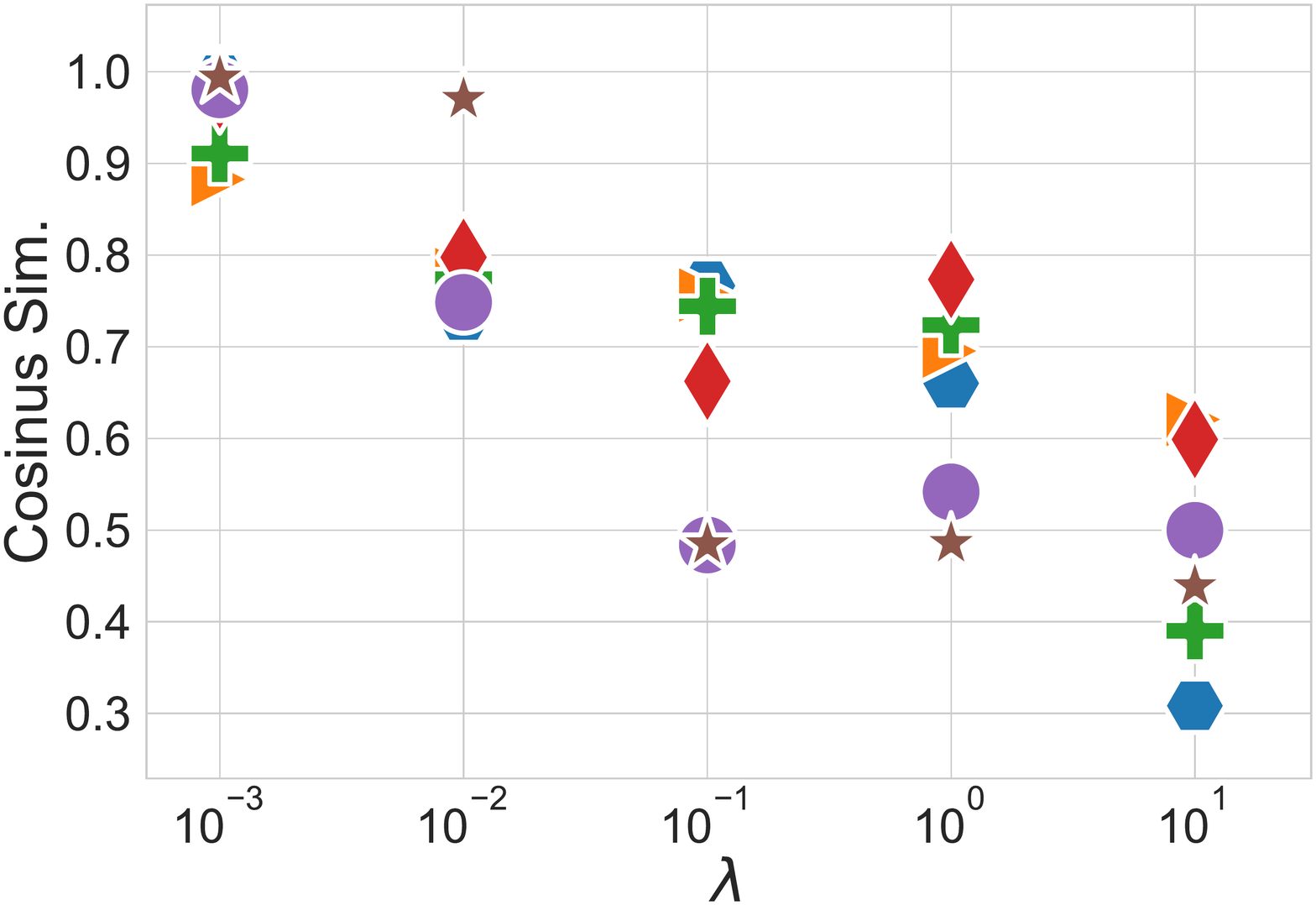}
         \caption{}\label{fig:coni_category_cg}
    \end{subfigure}
        \caption{Numerical experiments on the multi-class conditionnal sentence generation. Results include: BLEU (\autoref{fig:bleu_category_cg}); cosine similarity (\autoref{fig:coni_category_cg}); style transfer accuracy (\autoref{fig:accuracy_category_cg}); sentence fluency (\autoref{fig:ppl_category_cg}).}\label{fig:category_Cg}
\end{figure*}

\end{document}